\RequirePackage{fix-cm}
\documentclass[twocolumn]{svjour3}          % twocolumn
\smartqed  % flush right qed marks, e.g. at end of proof
\usepackage{graphicx}
\usepackage{bm}
\usepackage{amsmath}
\usepackage{amssymb}
\usepackage{marvosym}
\usepackage{epsfig}
\usepackage{graphicx}

\usepackage{amssymb}
\usepackage{tabu}
\usepackage{array}
\usepackage{algorithm}
\usepackage{algorithmic}
\usepackage{times}
\usepackage{epsfig}
\usepackage{graphicx}
\usepackage{amsmath}
\usepackage{multirow}
\usepackage{amssymb}
\usepackage{subfigure}
\usepackage{booktabs}       % professional-quality tables
\usepackage{amsfonts}       % blackboard math symbols
\usepackage{nicefrac}       % compact symbols for 1/2, etc.
\usepackage{microtype}      % microtypography
\usepackage{amsmath}
\usepackage{wrapfig}
\usepackage{amssymb}
\usepackage{color}

\global\long\def\vct#1{\boldsymbol{#1}}
\global\long\def\mat#1{\boldsymbol{#1}}

\global\long\def\optr{\mbox{tr}}
\global\long\def\opmat{\mbox{mat}}

\global\long\def\t#1{\widetilde{#1}}

\global\long\def\norm#1{\lVert#1\rVert}

\global\long\def\R{\mathbb{R}}

\global\long\def\vg{\boldsymbol{g}}

\global\long\def\vu{\boldsymbol{u}}

\global\long\def\vw{\boldsymbol{w}}
\global\long\def\vx{\boldsymbol{x}}

\global\long\def\vz{\boldsymbol{z}}

\newcommand{\tabincell}[2]{\begin{tabular}{@{}#1@{}}#2\end{tabular}}

\global\long\def\mA{\boldsymbol{A}}

\global\long\def\mC{\boldsymbol{C}}

\global\long\def\mM{\boldsymbol{M}}

\global\long\def\mP{\boldsymbol{P}}
\global\long\def\mQ{\boldsymbol{Q}}
\global\long\def\mR{\boldsymbol{R}}
\global\long\def\mS{\boldsymbol{S}}

\global\long\def\mV{\boldsymbol{V}}

\global\long\def\a{\alpha}

\global\long\def\T{\top}

\global\long\def\vu{\vct{u}}
\global\def\revised#1{ {#1} }
\definecolor{darkgreen}{rgb}{0,0.6,0}
\global\def\minorrevised#1{ {#1} }
\global\long\def\ie{\textit{i.e.}, }
\global\long\def\eg{\textit{e.g.}, }

\begin{document}

\title{Ensemble Quadratic Assignment Network for Graph Matching%\thanks{Grants or other notes
}


\author{
$\text{Haoru Tan}^{1,2}$         \and     $\text{Chuang Wang}^{1,2,\text{\Letter}}$   \and  $\text{Sitong Wu}^{2}$  \and  $\text{Xu-Yao Zhang}^{1,2}$ \and  $\text{Fei Yin}^{1,2}$  \and  $\text{Cheng-Lin Liu}^{1,2}$
}


\institute{ \textsuperscript{\rm 1}National Laboratory of Pattern Recognition, Institute of Automation of Chinese Academy of Sciences,  Beijing 100190, China\\
	\textsuperscript{\rm 2}School of Artificial Intelligence, University of Chinese Academy of Sciences, Beijing 100044, China\\
	\textsuperscript{\rm $\text{\Letter}$} Corresponding author: Chuang Wang (chuang.wang@nlpr.ia.ac.cn)
}

\date{Received: date / Accepted: date}

\maketitle

\begin{abstract}
Graph matching is a commonly used technique in computer vision and pattern recognition.
Recent data-driven approaches have improved the graph matching accuracy remarkably, whereas some traditional algorithm-based methods are more robust to feature noises, outlier nodes, and global transformation (e.g.~rotation).
In this paper, we propose a graph neural network (GNN) based approach to combine the advantage of data-driven and traditional methods. 
In the GNN framework, we transform traditional graph matching solvers as single-channel GNNs on the association graph and extend the single-channel architecture to the multi-channel network. 
The proposed model can be seen as an ensemble method that fuses multiple algorithms at every iteration.
Instead of averaging the estimates at the end of the ensemble, in our approach, the independent iterations of the ensembled algorithms exchange their information after each iteration via a 1x1 channel-wise convolution layer. 
Experiments show that our model improves the performance of traditional algorithms significantly.
In addition, we propose a random sampling strategy to reduce the computational complexity and GPU memory usage, so that the model is applicable to matching graphs with thousands of nodes.
We evaluate the performance of our method on three tasks: geometric graph matching, semantic feature matching, and few-shot 3D shape classification. The proposed model performs comparably or outperforms the best existing GNN-based methods.
\end{abstract}

\section{Introduction}
% background and define problem of GM
Graph matching (GM) is a fundamental technique in computer vision, widely used in object recognition~\cite{graph_rec1,graph_rec2}, tracking \cite{graph_track1,graph_track2,graph_track3}, and shape matching~\cite{graph_shape1,graph_shape2,graph_shape3} tasks.
Given two graphs with certain node features, e.g. locations of feature points~\cite{DGFL}, GM
 aims to find the correspondence between the two sets of nodes, which is a typical combinatorial optimization problem.

 % traditional method
In traditional algorithm-based approaches, GM is usually decomposed into two stages: modeling and optimization. At the modeling stage, the node features and graph structures are used to construct an affinity matrix, which measures the similarity  of the nodes and edges from the two graphs. At the optimization stage,  GM is formulated as a quadratic assignment problem (QAP) to find a 0-1 assignment vector that maximizes the sum of the affinity scores of all matched nodes and edges. Since QAP is an NP-hard problem, researchers 
developed a variety of relaxation strategies for efficiently approximating the solution of this optimization problem \cite{sp1,relax1,relax2}.

% recent learning based method
Recently, benefiting from the deep learning technology, a series of data-driven approaches have been proposed to  treat the two steps of  feature representation learning and combinatorial optimization jointly~\cite{pia,CIE,blackbox,DGMC,DGFL,LCSGM,NGM}, and have achieved remarkable  improvement in matching accuracy on several public datasets. 
The improvements largely owe to learning better feature representations, while the optimization solver is mostly akin to the traditional algorithms~\cite{gmn,AAAI2021,SIGMA,blackbox,DLGM}. 
The integration of the optimization solver into an end-to-end trainable model is done by transforming the affinity matrix into a differentiable map.
For example, the power method for solving the leading eigenvector of the affinity matrix corresponds to the recent work, 
GMN~\cite{gmn}, and the proximal method for solving a relaxed QAP problem relates to the proximal matching neural network~\cite{AAAI2021}. 
Further, Rolinek~et~al.~\cite{blackbox} showed that many other traditional methods can also be used to build a black-box neural network for graph matching.

The data-driven models, despite their high prediction accuracy of node correspondence on given datasets, often suffer from data distribution shifts~\cite{DGFL}, e.g., random rotation. 
On the contrary, traditional methods are more robust to data transformations and noises. 

In this paper, we propose an ensemble quadratic assignment network (EQAN) that combines the advantages of both the traditional and data-driven methods.
Our model is constructed based on an observation that a traditional algorithm corresponds to a single-channel GNN model on an association graph, which is the Cartesian product of the two input graphs, and therefore a multi-channel GNN on the association graph relates to an ensemble of  multiple base solvers.
Instead of naive average of the multiple independent algorithms at the end of the iterations, in our approach, the base solvers exchange their information right after each iteration via a $1\times 1$ channel-wise convolution layer. Moreover, we introduce an affinity update scheme to further boost the matching accuracy.
Experiments show that the proposed model improves traditional methods dramatically, whereas the naive ensemble method does not yield noticeable change. 

%Additional random sampling method
In addition, to handle large graphs with thousands of nodes, we present a random sampling strategy, which reduces the GPU memory footprint and the theoretical computational complexity of EQAN. 
Experiments show that the random sampling strategy improves the scalability of EQAN dramatically without significant performance loss.
This strategy allows us to apply the proposed method  to applications requiring a huge number of nodes, for example, few-shot 3D shape classification.

We evaluate the performance of our method on three tasks: geometric graph matching, semantic feature matching, and few-shot 3D shape classification. The proposed model performs comparably or outperforms the best existing GNN-based methods.

\revised{
This work is built up on our conference paper \cite{AAAI2021} with extensively  improvement and new contributions. In  \cite{AAAI2021}, we developed a differentiable proximal matching (DPGM) network for graph matching based on the proximal optimization method, which can be considered as a single-channel GNN. In this work, we design a multi-channel deep ensemble network, where each channel is a single-channel QAP-solver, \eg proximal optimizer. However, the base matching solver is not limited to the proximal method. Other traditional methods, for example, the spectrum method \cite{sp1}, the graduated assignment method \cite{GAGM} and the re-weighted random walk matching (RRWM)  \cite{RRWM}, can also be integrated into the proposed framework, and gain a significant performance improvement as well. Experiments show that we achieved an improvement of approximately 5.6 percent on geometric graph matching tasks and 13.0 percent on semantic feature matching tasks compared with our preview work \cite{AAAI2021}. 
}

%	This work is built up on our conference work, which proposed the differentiable proximal graph matching  (DPGM) network~\cite{AAAI2021}. In this work,  DPGM  is used as a component QAP solver of the ensemble network.
%	 Meanwhile, the proposed ensemble framework is also compatible to other graph matchers. For example, one can deploy  the spectrum method~\cite{gmn} or the graduated assignment method~\cite{GAGM} as the component of QAP solvers as well. The improvements of ensemble network with the alternative embedded QAP solvers are also demonstrated in the experiments.

The main contributions of this paper are summarized as follows.
\begin{itemize}
\item We introduce the ensemble quadratic assignment network (EQAN) for solving graph matching problems. It combines the advantages of traditional and data-driven methods to enhance both matching accuracy and robustness.
\item We proposed a random sampling strategy to improve the scalability so that our method can fit problems with thousands of nodes.
\item Experiments on three computer vision tasks, geometric graph matching, semantic feature matching, and few-shot 3D shape classification, reveal the good performance of our model.
\end{itemize}

	The remaining paper is organized as follows. Section 2 reviews related works. Section 3 introduces the graph matching task and differentiable solvers. Section 4 describes our ensemble framework.	Section 5 formulates our method as a multi-channel graph neural network, and introduces the affinity update scheme to boost the performance and the random sampling strategy for better scalability. Section 6 presents the experimental results, and Section 7  draws concluding remarks.

\section{Related Works}

	\subsection{Traditional graph matching algorithms}

	Traditional works developed varieties of relaxation and approximated optimization strategies for solving the optimization problem of graph matching (GM) tasks. A series of GM algorithms~\cite{sp1,sp2,sp3,sp5} were built based on the spectral technique. They converted the GM problem to the task of finding the leading eigenvector of the affinity matrix.
	%The solving process of this method is often very efficient, but it also has the disadvantages of inferior performance and poor robustness to noise.
	%The bi-stochastic relaxation technique (also known as the doubly stochastic relaxation) replaces the space of permutations with the space of doubly stochastic matrices.
	GAGM~\cite{GAGM} uses a gradient-based graduated assignment technique under the bi-stochastic relaxation technique (also known as the doubly stochastic relaxation). This  relaxation replaces the space of permutation matrices with  the space of doubly stochastic matrices.
	RRWM~\cite{RRWM} is a random-walk-based method that transforms the GM  into  the problem of selecting reliable nodes from the association graph.
	IPFP~\cite{IPFP} directly computes the solution in the discrete assignment space without the  continuous-variable relaxation.

	Usually, the plain implementations of the GM algorithm have high computational complexity.  A line of works aims to improve the scalability by avoiding maintaining the complete affinity matrix. For instance, FGM~\cite{fgm} decomposes the affinity matrix into smaller blocks and introduces global geometric constraints.	KerGM~\cite{KerGM} provides a unified perspective for Koopmans-Beckmann’s and Lawler's QAP by introducing new rules for array operations in the associated Hilbert spaces. Wang~et~al.~\cite{functional} proposed a functional representation for graph matching to avoid computing the affinity matrix.

	The traditional approaches primarily focus on developing better and faster algorithms for solving the underlying optimization problem. In those works, the graph features and affinity matrix are mostly hand-crafted. In practice, seeking a good representation of node features greatly affects the performance~\cite{gmn}.

\subsection{Data-driven graph matching models}

    Depending on the training data type and the input feature, we roughly classify the existing data-driven methods into two types. The first type focuses on visual feature matching~\cite{DGMC,pia,gmn,blackbox,shaof2022m}, in which pixels information are available and the models usually incorporate a deep convolutional neural network (CNN), {\em e.g.}, VGG~\cite{vgg} for feature encoding. The second type is referred to as geometric matching, where only the locations of nodes are given. In this case, a graph neural network (GNN)~\cite{DGFL,DGMC,Pointnet,AAAI2021} is deployed to generate high-order features by aggregating local geometric information. Other methods~\cite{DGMC,LCSGM}  are applicable to both situations.

	Recent data-driven approaches combine feature extractor and optimization solver in a unified model and train the model in an end-to-end fashion.
	Wang~et~al.~\cite{pia} proposed a differentiable deep network pipeline for learning soft permutation and graph embedding.
	Yu~et~al.~\cite{CIE} proposed a deep graph matching learning method with channel-independent embedding (CIE) and Hungarian attention.
	The above GM algorithms involve the Hungarian method to discretize the continuous solution, in which the discretization procedure cuts off the  end-to-end computational graph,   Rolínek~et~al.~\cite{blackbox} addressed this challenge by estimating the gradient flow of the matching algorithm in a black-box manner.
 Fey~et~al.~\cite{DGMC} proposed a two-stage graph neural matching method, called deep graph matching consensus (DGMC), that achieves remarkable performance by incorporating a local consensus requirement.
	Gao~et~al.~\cite{QCDGM} presented a deep learning framework that explicitly formulates pairwise graph structures as a quadratic constraint.
	Zeng~et~al.~\cite{ZENG20218} proposed a parameterized graph matching framework by linearly combining multiple Koopmans–Beckmann’s graph matchings.
	Liu~et~al.~\cite{reinforcementGM} proposed a reinforcement learning approach for sequentially finding the node-to-node matching.
	Liu~et~al.~\cite{SIGMA} proposed a stochastic iterative graph matching model (SIGMA) by defining a distribution of matchings for the input graph pair.
	Yu~et~al.~\cite{DLGM} proposed a deep latent graph matching model (DLGM) to  learn the matching results and the latent topology of the input graph jointly.

	A few works have involved  the association graph for graph matching~\cite{RRWM,LCSGM,NGM}. Cho~et~al.~\cite{RRWM} showed that computing the node-wise correspondence is equivalent to picking trustworthy nodes from the association graph. Wang~et~al.~\cite{LCSGM} proposed a novel neural combinatorial optimization solver on the association graph, which seeks to learn affinity and solvers simultaneously.	Wang~et~al.~\cite{NGM} expands the GNN solver (NGM) on the association graph applicable to more broad QAP problems.

	On  data utilization, the deep graphical feature learning (DGFL) method \cite{DGFL} introduced the synthetically supervised pipeline, in which the model is trained with only synthetic random graphs and is tested on real-world datasets.
	Wang~et~al.~\cite{NeurIPS_Wang} presented an unsupervised graph matching technique that simultaneously learns graph clustering and graph matching.

\revised{
Though existing GNN-based  approaches have achieved remarkable improvement in term of matching accuracy on several public datasets, they are usually not robust to data transformations and noises~\cite{DGFL} as  traditional optimization-based methods~\cite{AAAI2021}. Figure \ref{fig:unrobust deepGM} demonstrates a concrete comparative experiment.
The proposed EQAN not only inherits the good robustness of traditional methods but also maintains the high accuracy of deep GM methods via mapping
 an ensemble of several base solvers (not limited to our DPGM) as a multi-channel GNN with additional designation on how to share information among channels and how to utilize the affinity matrix.
}

\subsection{Applications}

	The surveys \cite{ICMR,survey_1,survey_2} reviewed application scenarios of graph matching methods. Here are some typical  examples of computer vision and pattern recognition. Shen~et~al.~\cite{app_image_reg} applied graph matching tools to medical image registration. 	Wu~et~al.~\cite{app_new_remote_1} presented a novel graph matching approach for remote sensing image registration.	Nie~et~al.~\cite{app_object_tracking} and He~et~al.~\cite{app_new_track_1} use graph matching tools to solve the problem of object tracking.	Shen~et~al.~\cite{app_ReID}, Wang~et~al.~\cite{app_new_reid_1}, and Zhao~et~al.~\cite{app_new_reid_2} proposed  using graph matching methods to tackle the person re-identification problem. Wang~et~al.~\cite{app_scene_understanding} proposed a scene-understanding approach based on graph matching.
	Liao~et~al.~\cite{app_protein} and Zhang~et~al.~\cite{KerGM} applied graph matching methods to the problem of aligning multiple protein networks.	Fiori~et~al.~\cite{app_mri} used graph matching algorithms to infer brain connectivity from functional magnetic resonance imaging data.	He~et~al.~\cite{app_new_semantic_1} proposed a GM-based approach to find the semantic correspondences between images.	Justin~et~al.~\cite{proximal_gm2} and Fu~et~al.~\cite{app_new_point_1} used graph matching tools to analyze 3D CAD models and point clouds.
	Chopin~et~al.~\cite{ssgm} reformulated image segmentation as an inexact graph-matching problem with many-to-one constraints. 
 In addition to these, many practical tasks, such as action recognition \cite{app_action_recog}, visual localization \cite{app_new_visual_localization_1}, real-time matching of image wireframes \cite{app_new_wireframes_1}, pose estimation \cite{liujh2023prior}, etc., also need to estimate the point-to-point correspondence between multiple images. 

In all these applications, high matching accuracy and efficiency (scalability to large graphs) are always desirable.

\section{Graph Matching and Differentiable Solver}	\label{sec:pre}

\subsection{Problem formulation}

%\subsubsection*{Problem formulation}
Graph matching aims to find the correspondence of nodes between two graphs. Let ${G}_1 = (V_1, E_1)$ and ${G}_2 = (V_2, E_2)$ be two undirected graphs. We assume that the graphs have an equal number of nodes $|V_1|=|V_2|=n$. For the unequal case, one can always add additional dummy nodes to meet this condition.

Formally, graph matching can be formulated as a quadratic assignment problem (QAP) \cite{sp1}
\begin{equation} \label{eq:opt}
\begin{aligned}
\max_{\vx  \in \{0,1\}^{n^2}} & \vx^\T \mM \vx \quad \text{s.t.} &\quad \bm{\mathcal{C}} \vx = \mat{1},
\end{aligned}
\end{equation}
where
$\bm{x}$ is an assignment vector such that $[\vx]_{ij}=1$ if  node $i$ in ${G}_1$ corresponds to node $j$ in ${G}_2$, and 0 otherwise; the {affinity matrix} $\mM \in \mathcal{R}_{+}^{n^2 \times n^2}$ measures the similarity between the nodes in the two graphs;
the binary matrix $\bm{\mathcal{C}}$ encodes $2n$ linear constraints:
$\sum_{i\in V_1} [{\vx}]_{ij^\prime} = \sum_{j \in V_2} {[\vx]}_{i^\prime j}=1$ for all $i^\prime \in V_1$ and $j^\prime \in V_2$. These constraints  ensure that the matching map induced by any valid $\vx$ is bijective from $V_1$ to $V_2$.

The affinity matrix is constructed based on the topology of the graphs and the local features associated with the nodes and edges.
The diagonal terms $[\mM]_{ii,jj}$ represents the reward if node $i$ in ${G}_1$ is mapped to node $j$ in ${G}_2$. The off-diagonal terms $[\mM]_{ii^\prime,j j^\prime}$ represents the reward if a pair of nodes $(i,i^\prime)$ in ${G}_1$  corresponds to a pair $(j,j^\prime)$ in ${G}_2$.
A classical example is Koopman-Beckmann's construction $\mM = \mA_1 \otimes \mA_2$ \cite{koopman}, where $\mA_1$ and $\mA_2$ are adjacent matrices of ${G}_1$ and ${G}_2$ respectively. Intuitively, $[\mM]_{ii^\prime, jj^\prime} = 1$ if and only if both pairs of $(i, i^\prime)$ and $(j, j^\prime)$ are directly connected in ${G}_1$ and ${G}_2$ respectively, and   $0$ otherwise. %This construction only uses the topology structure of the two graphs without any other information.

\subsection{Differentiable proximal graph matching }
\label{sec:dpgm}
%\subsubsection{Derivation}
We introduce a classical algorithm to solve \eqref{eq:opt} approximately based on the proximal method \cite{Proximal_method}. In general, the integer optimization problem \eqref{eq:opt}  is hard to solve. First, we relax it to a continuous optimization problem
\begin{equation} \label{eq:opt2}
\begin{aligned}
\min_{\vz \in \mathcal{R}_{+}^{n^2}} \;  \mathcal{L}(\vz) =   - \vz^\T \mM \vz  -  h(\vz)\;\;\;
\text{s.t.} \;\; \bm{\mathcal{C}} \vz = \mat{1},
\end{aligned}
\end{equation}
where {$h(\vz) = - \lambda  \vz^\T \log(\vz)$} is an entropy regularizer with the scalar $\lambda$ being the regularizer parameter. The matrix $\bm{\mathcal{C}}$ is the same as the one in \eqref{eq:opt}. The parameter $\lambda$ controls the hardness of the optimization problem \eqref{eq:opt2}.

\begin{algorithm}[tp]
	\caption{:~~Differentiable Proximal Graph Matching\label{alg:DPGM}}
	\begin{algorithmic}
		\STATE {\bfseries Input:} Affinity matrix $\mM$, maximum iteration $K$ and stepsize $\beta$.
		\STATE {\bfseries Initialization:} $\vz_0 \Longleftarrow {\rm Sinkhorn}(\vu)$.
		\FOR{$k=1$ {\bfseries to} $K$}
		\STATE $\widetilde{\vz}_{k} \Longleftarrow
\exp\Big[
\tfrac{\beta}{ 1+ \lambda \beta}  ( \mM \vx_{k-1})
+ \tfrac{1}{1+\lambda \beta} {\log(\vz_{k-1})}
\Big]$
		\STATE $\vz_{k} \Longleftarrow  {\rm Sinkhorn}(\widetilde{\vz}_{k})$
		\ENDFOR
		\STATE Output $\vz_{K}$
	\end{algorithmic}
\end{algorithm}

\begin{algorithm}[tp]
		%\large
		\caption{:~~Sinkhorn Method \label{alg:sinkhorn}    }
		\begin{algorithmic}
			\STATE {\bfseries Input:}  $\bm{X} \in \mathcal{R}_{+}^{n \times n}$,  $T=5$.
			\STATE {\bfseries Initialization:}  $\bm{b} \leftarrow \bm{1}, \mu \leftarrow \bm{1}^{n \times 1}, \bm{v} \leftarrow \bm{1}^{n \times 1}$.
			\FOR{$t=0$ {\bfseries to} $T-1$}
			\STATE \quad $\bm{a} \leftarrow \frac{\mu}{\bm{\bm{X} b}}$,  $\bm{b} \leftarrow \frac{\bm{v}}{\bm{X}^{{\T}}\bm{a}}$.
			\ENDFOR
			\STATE $\bm{Y} = \text{Diag}(\bm{a})\bm{X} \text{Diag}(\bm{b})$
			\STATE {\bfseries Output:} $\bm{Y}$
		\end{algorithmic}
\end{algorithm}

Then, we decompose the non-convex objective \eqref{eq:opt2} into a sequence of convex sub-problems
    \begin{equation} \label{eq:prox}
    	\bm{z}_{t+1} =   \mathop{\rm argmin}_{\bm{z} \in \mathcal{R}_{+}^{n^2 \times 1}, \bm{\mathcal{C}} \vz = \bm{1}}\;
    	\bm{z}^{\T} \nabla g(\bm{z}_t) - h(\bm{z}) + \frac{1}{\beta}
    	D(\bm{z}, \bm{z}_t)
    \end{equation}
    where $g(\bm{z})= -  \vz^\T \mM \vz$; the function $D(\cdot, \cdot)$ is a non-negative proximal operator with  $D(\bm{x}, \bm{y}) = 0$ if and only if $\bm{x} = \bm{y}$; and $\beta>0$ is a scalar parameter controlling the proximal weight.
    Since $[\vz]_{ij} \in [0,1]$, we can interpret it as the probability of mapping node $i$ in ${G}_1$ to node $j$ in ${G}_2$.
    We choose the Kullback-Leibler divergence as the proximal function, \ie  $\;D(\vz, \t{\vz})=  \vz^\T\log(\vz) -\vz^\T \log(\t{\vz})$.

    The convex optimization \eqref{eq:prox} has an analytical solution
\begin{align} \label{eq:iter-1}
\widetilde{\vz}_{t+1} &=
\exp\Big[
{ w_p  \mM \vz_t } + w_z \log(\vz_t)
 \Big],
\\ \label{eq:iter-2}
\vz_{t+1} & = {\text{Sinkhorn}(\widetilde{\vz}_{t+1})},
\end{align}
where $w_p = \tfrac{\beta}{ 1+ \lambda \beta}$, $w_z = \tfrac{1}{1+\lambda \beta} $, and Sinkhorn$(\cdot)$ is the Sinkhorn-Knopp transform. \revised{The proof of convergence of the above DPGM algorithm and the detailed derivation of \eqref{eq:iter-1}, \eqref{eq:iter-2} from \eqref{eq:prox}  are presented in Appendix A.2 and A.3 respectively.} Pseudo-codes are shown in Algorithm \ref{alg:DPGM} and Algorithm \ref{alg:sinkhorn}. When applying the Sinkhorn-Knopp transform, we reshape the input (output) as a $n\times n$ matrix ($n^2$-dimensional vector) respectively. One can check that \eqref{eq:iter-1} solves \eqref{eq:prox} ignoring the constraint  $\bm{\mathcal{C}} \vz = \mat{1}$, and the normalization step \eqref{eq:iter-2} ensures that $\opmat(\vz_{t+1})$ is a $n\times n$ doubly stochastic matrix.

\subsection{General differentiable iterative solver}

The proximal method  introduced above is one of a large set of classical algorithms for solving the quadratic assignment problem (QAP) \eqref{eq:opt}. Other methods are, for example, the spectrum method (SM) \cite{gmn}, and the graduated assignment graph matching (GAGM) algorithm \cite{GAGM}.
The iteration equation of all these methods can be abstracted as a single transform
\begin{equation} \label{eq:matcher-or}
		\vz_{t+1}=  ~\text{QAP-Solver}\Big(  \vz_{t},\bm{M}; \vw_t\Big),
\end{equation}
where $\vw_t$ encloses all algorithmic parameters at step $t$. In the proximal method \eqref{eq:iter-1}, \eqref{eq:iter-2}, the parameters are $\vw_t=[w_p,w_z]$.
The spectrum method is a power iteration for solving the eigenvector of the leading eigenvalue of the  affinity matrix, \ie $\;\vz_{t+1} = \mM \vz_{t} / \norm{\mM \vz_{t}}$. We listed the details of SM and GAGM methods in Appendix B.

Recent data-driven approaches \cite{gmn,AAAI2021,blackbox} are based on an observation that the update transforms \eqref{eq:matcher-or} is a differentiable map from $\mM$ and $\vz_t$ to $\vz_{t+1}$. Therefore, one can integrate the graph matching algorithm into a deep-learning framework by combing the feature extractor and the matching algorithm together and training them in an end-to-end manner. For example, GMN \cite{gmn} used the spectrum method; our previous work \cite{AAAI2021}, DPGM deployed the proximal method; and BBGM \cite{blackbox} considered \eqref{eq:matcher-or} as a general black-box solver.

\revised{ \subsection{QAP-Solver as a single-channel GNN }}
\label{sec:QAP-Solver as a single-channel GNN}
In what follows, we show that a differentiable QAP-solver can be considered as a single-channel GNN on an association graph of two base graphs $G_1=(V_1,E_1)$ and $G_2=(V_1, E_2)$.

\begin{figure}[tp]
	\centering
	\includegraphics[width=0.99\linewidth]{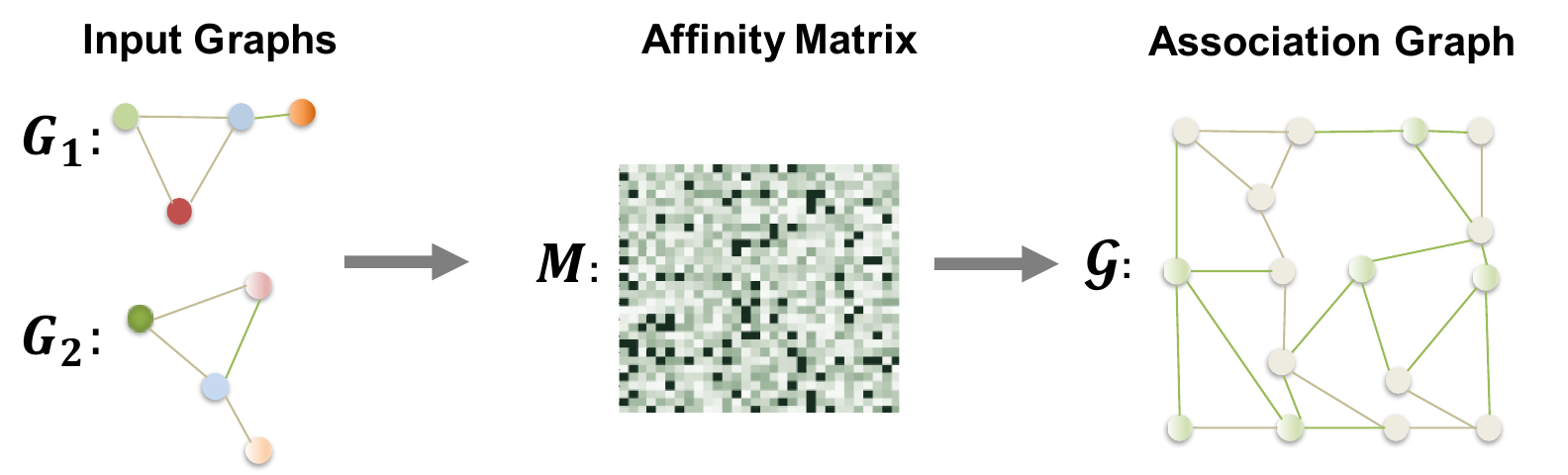}
	\caption{ \label{association_graph} The association graph $\mathcal{G}=(\mathcal{V}, \mathcal{E})$ of two base graphs $G_1=(V_1,E_1)$ and $G_2=({V}_2, {E}_2)$. Each association-node from the association graph represents a candidate matching between a pair of nodes from two base graphs. Graph matching is equivalent to selecting a set of nodes in the association graph \cite{RRWM,LCSGM,NGM} with certain exclusive constraints.
	}
\end{figure}

The association graph $\mathcal{G}=(\mathcal{V}, \mathcal{E})$, with an example shown in Figure~\ref{association_graph}, is constructed as follows.
The association-node set $\mathcal{V}$ is the Cartesian product of the two base node sets, i.e., $\mathcal{V} = V_1 \otimes V_2 = \{ i j | i \in V_1, j \in V_2 \} $.
 %(To avoid confusion, we refer to the node on an association graph as a association-node.)
Each association-node indexed by the tuple $ij$ from the association graph represents a candidate matching between a pair of nodes $\{i \in V_1, j \in V_2 \} $ from two base graphs.
The association-nodes $ij$ and $i^\prime j^\prime$ from the association graph are connected if and only if $(i,i^\prime)$ and $(j,j^\prime)$ are edges in $E_1$ and $E_2$.

To illustrate the connection between a QAP-solver and GNN, we take the DPGM method \eqref{eq:iter-1}, \eqref{eq:iter-2} as a concrete example.
In particular, the matrix multiplication $\mM \bm{z}_\text{t}$ in \eqref{eq:iter-1} can be decomposed into two neural message passing steps \cite{Message_Passing}: the aggregation step and the local update step
\begin{equation}
\begin{aligned} \label{eq:update-rule-overall}
 \Big[\widetilde{\bm{z}}_{t+1}\Big]_{ij} = \exp\Big[  \underbrace{ w_p \sum_{i^{\prime} j^{\prime} \in \mathcal{N}_{ij} } [\mM]_{i i^{\prime}, j j^{\prime}}   [\vz_t]_{i^{\prime}j^{\prime}} }_\text{aggregation}
\\
+ \underbrace{ w_p [\mM]_{ii, jj} [\vz_t]_{ij}}_\text{local update} + \underbrace{  {w_z \log([\vz_t]_{ij}) }  }_\text{Damping} \Big].
\end{aligned}
\end{equation}
%\cw{check whether $[\mM]_{ii, jj}$ should be $[\mM]_{ij, ij} $. If yes, change the following text accordingly.}
In the aggregation step, the pairwise affinity $[\mM]_{i i^\prime, j j^\prime}$ is the weight to aggregate the neighbor's message $[\bm{z}_t]_{i^\prime j^\prime}$ to the central association-node $ij$.
In the local update step, the central association-node $ij$ update its previous value $[\bm{z}_{t}]_{ij}$ according to the unary potential $[\mM]_{ii, jj}$.
The exponential function acts as the nonlinear activation unit and $w_p$ is used as the transformation parameter during message passing.
Then, the Sinkhorn operation \eqref{eq:iter-2} normalizes  $\widetilde{\bm{z}}_{t+1}$ and produces the output ${\bm{z}}_{t+1}$.

The $ij$th association-node message $[\bm{z}_{t}]_{ij}$  in the above message passing rule is a scalar.
Therefore, the corresponding GNN only has a single channel, which usually has a limited capacity for modeling complex relationships.
Nevertheless, this analogy motivates us to develop a multi-channel GNN on the same association graph resulting in the proposed ensemble model, in which the feature  of the $ij$th association-node is generated by fusing the corresponding association-node messages   from multiple QAP-solvers  (single-channel GNNs). Details are illustrated in the following two sections.

\vspace{1em}

\section{Ensemble Quadratic Assignment Model}

\begin{figure*}[tp]
		\centering
		\includegraphics[width=1\linewidth]{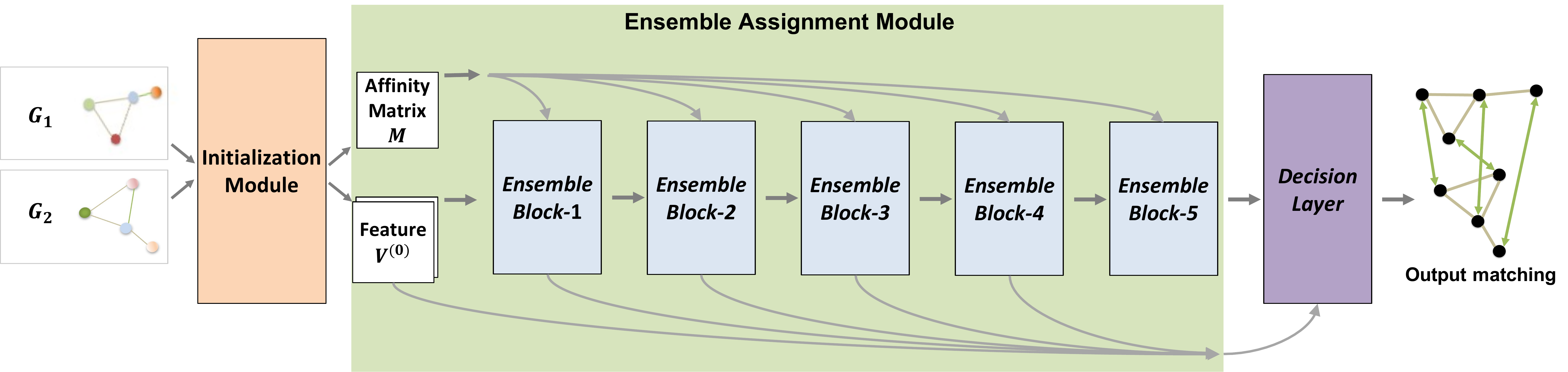}
		\caption{The ensemble quadratic assignment model consists of three modules. The initialization module computes the affinity matrix and initializes the iteration feature tensor. The ensemble assignment module is a sequence of ensemble blocks, which include a set of classical QAP solvers. Details are shown in Figure~\ref{flow_chart}. The final layer takes the features created by all previous blocks as input and outputs a matching decision.}
		\label{overview}
\end{figure*}

\begin{figure}[tp]
		\centering
		\includegraphics[width=0.99\linewidth]{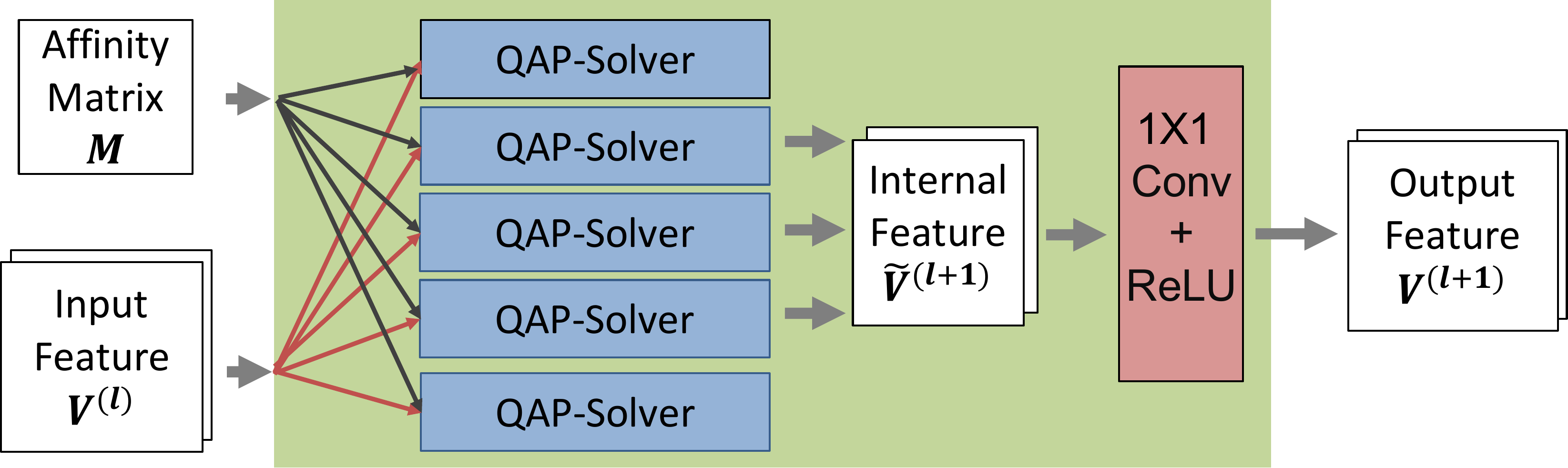}
		\caption{An \textit{Ensemble Block} includes a set of  QAP-solvers that update the channels of feature independently, and a $1\times 1$ convolution layer that promotes information exchange across  the channels.}
		\label{flow_chart}
\end{figure}

	As depicted in Figure \ref{overview}, the proposed ensemble quadratic assignment model consists of three sequential modules. 
	First, the  \textit{initialization module} constructs initial node features from raw inputs. 
	Then, the \textit{ensemble assignment module} refines the node features using a sequence of ensemble blocks. Each block is a set of QAP-solvers. 
	Finally, the \textit{decision layer} generates the final matching score and prediction from the refined features.

\subsection{Initialization module}
	\label{sec:initialization_module}
	
	 The initialization module aims to construct the affinity matrix $\mM$ and initialize a feature tensor $\mV^{(0)}$ from the raw inputs, e.g. location of nodes and local pixel values. This module can be formulated as 
	 \begin{equation} \label{eq:initialization_module}
		{\mV}^{(0)}, \mM =  ~\text{Initialization-Module}\Big(  G_1, G_2 \Big),
	\end{equation}
where $G_1$ and $G_2$ contain the raw node features and graph structures of input graphs respectively. Inside this module, firstly, we convert the raw input features associated with the nodes of each graph  into a set of $d$-dimensional vectors denoted by $  \bm{F}_q\in \mathcal{R}^{ n \times d}$, $q=1,2$.
	 A vanilla implementation is simply concatenating all raw information with proper value scaling.  
	 A better but optional method is employing a graph neural network (GNN) to extract local features by aggregating a certain range of neighbor information \cite{DGFL,DGMC,pia}.

	 %The diagonal terms $[\mM]_{ii,jj}$ represents the reward if node $i$ in ${G}_1$ is mapped to node $j$ in ${G}_2$. The off-diagonal terms $[\mM]_{ii^\prime,j j^\prime}$ represents the reward if pair $(i,i^\prime)$ in ${G}_1$  corresponds to pair $(j,j^\prime)$ in ${G}_2$.

	 Then, we construct the affinity matrix $\bm{M}$ by setting the diagonal terms $[\mM]_{ii, jj}= \norm{ [\bm{F}_1]_{i,:} - [\bm{F}_2]_{j,:} }$, and  the off-diagonal terms
	 \begin{equation} \nonumber%\label{stem:m}
	 	\begin{aligned}&
	 		[\mM]_{ii^\prime,j j^\prime} =
	 		\begin{cases}
	 			\text{e}^{ -|d^{(1)}_{i i^\prime}-d^{(2)}_{j j^\prime }|^2/\sigma^2 }, \text{ if } (i, i^\prime) \in G_1,\;  (j, j^\prime) \in G_2, \\
	 			0, \text{ otherwise },
	 		\end{cases}
	 	\end{aligned}
	 \end{equation}
	 where $d^{(1)}_{ii^\prime}=\norm{ [\bm{F}_1]_{i,:} - [\bm{F}_1]_{i^\prime,:} }$ and $d^{(2)}_{jj^\prime}=\norm{ [\bm{F}_2]_{j,:} - [\bm{F}_2]_{j^\prime,:} }$, and $\sigma$ is a (predefined or learnable)  parameter.

	 Lastly, we initialize the multi-channel feature tensor $\bm{V}^{(0)}$ by concatenating the raw features and performing a channel-wise linear transform. Specifically, the operation is
	 \begin{align}
	 &[\hat{\bm{V}}]_{:,i,j} = \text{Concate}\Big(  | [\bm{F}_1]_{i,:} - [\bm{F}_2]_{j,:} |,  [\bm{F}_1]_{i,:},  [\bm{F}_2]_{j,: }\Big),%\nonumber
	 \\
	 \label{eq:init_relu}
	 &\bm{V}^{(0)} = \text{ReLU} \Big( \text{Conv-1$\times$1} ( \hat{\bm{V}} ) \Big),
	 \end{align}
	 where $| \cdot |$ is the entry-wise absolution operation, and the operation Conv-$1\times 1( \hat{\bm{V}})$ is the $1\times 1$ convolutional operation mapping from $\mathcal{R}^{(3d)\times n\times n}$ to  $\mathcal{R}^{C\times n\times n}$ by performing a linear transform along $\hat{\bm{V}}$'s first coordinate axis (enumerating the indices of the second and third axes).

\subsection{Ensemble assignment module}

The ensemble assignment module refines the feature tensor $\mV^{(\ell)}$ iteratively, $\ell =1,2,\ldots, L$. Formally, we write
\begin{align}  \label{eq:block}
		&\bm{V}^{(\ell)}   = \text{EnsembleBlock}^{(\ell)} \Big(  \bm{V}^{(\ell-1)} ~;~\bm{M} \Big),
		%\\ \label{eq:concat}
		%&\bm{V} = \text{Concate}([\bm{V}^{(0)},  ...\bm{V}^{(L)} ] )\in \mathcal{R}^{(L+1) C \times n_1 \times n_2},
\end{align}
%hierarchically
with the initial feature tensor $\bm{V}^{(0)}$ and $\bm{M}$ received from the initialization module.
In this module, all {\em EnsembleBlock } use the same affinity matrix $\bm{M}$.

%We acquire the global decision feature $\bm{V}$ as the module output by concatenating the features from all blocks.

The {\em EnsembleBlock }
 consists of a set of parallel QAP solvers \eqref{eq:matcher} and a $1\times 1$ convolution layer.
We draw its detailed structure in Figure \ref{flow_chart}.
The QAP-solvers update the feature tensors  channel-by-channel independently as a QAP layer, {\em i.e.}
\begin{equation} \label{eq:matcher}
		[\widetilde{\mV}^{(\ell)}]_{c,:,:}=  ~\text{QAP-Layer}\Big(  [\mV^{(\ell-1)}]_{c,:,:} + \epsilon, \bm{M}; \vw^{(\ell,c)}\Big),
	\end{equation}
where $[\mV^{(\ell-1)}]_{c,:,:}\in \mathcal{R}^{n \times n}$ is the slice of the feature tensor with a given channel $c$; $ \vw^{(\ell,c)}$ is the internal parameters inside the solver; $\epsilon$ is a very small number set as $10^{-5}$ to avoid the error,  $\log 0 = -\text{inf}$, when ReLU = 0 in \eqref{eq:init_relu} and \eqref{eq:conv}.
The QAP-Layer  can be any classical matching algorithm, as long as the corresponding map is differentiable.

	After the channel-wise  update, we use the $1\times1$ convolution layer to exchange information of $\widetilde{\mV}$ among channels
	\begin{equation} \label{eq:conv}
	\mV^{(\ell)}= \text{ReLU}\Big[ \text{Conv-}1\times1\big(  \widetilde{\mV}^{(\ell)}\big)  \Big].
	\end{equation}
	By combining \eqref{eq:matcher} and \eqref{eq:conv}, we build the  update  equation  \eqref{eq:block} for a single EnsembleBlock.

	Finally, after $L$ iterations of \eqref{eq:block}, we concatenate all feature tensors $\{\mV^{(\ell)}\}$ as the output of this module,
	\begin{equation} \label{eq:concat}
	\bm{V} = \text{Concate}([\bm{V}^{(0)},  ...\bm{V}^{(L)} ] )\in \mathcal{R}^{(L+1) C \times n \times n}.
	\end{equation}
	Experiments show that  fusing of all feature tensors performs better than only using the last one.

\subsection{The decision layer}

The decision layer maps the global feature $\mV$ defined in \eqref{eq:concat} to a reward matrix $\bm{R} \in \mathcal{R}_{+}^{n \times n}$ via
	\begin{equation} \label{reward_matrix}
			\bm{R} = \exp \Big(   \text{Conv-1$\times$1}(  \bm{V}   )     \Big),
	\end{equation}
	where the $1\times 1$ convolutional operation performs the linear transform along $\mV$'s first coordinate axis. The entry $[\bm{R}]_{i j}$ is a positive reward of matching node $i $ in $ G_1$ with node $j $ in $G_2$.  Then, we normalize $\bm{R}$ by
	\begin{equation}   \label{matching_score}
			\bm{Q} =  \text{Sinkhorn}\Big(  \bm{R} \Big)
	\end{equation}
	and get a non-negative soft matching prediction matrix $\bm{Q}$.
	The entry $[\bm{Q}]_{ij}$ can be interpreted as the estimated probability that node $i$ in $G_1$ matches node $j$ in $G_2$.

	In the training phase, we use the cross entropy loss \cite{pia} to calculate the loss between the prediction matrix $\bm{Q}$ and the ground-truth assignment $\bm{x}^\ast$,
	\begin{equation}  \label{eq:loss}
		\begin{aligned}
			L (\mQ, \vx^\ast) = - \langle   \vx^\ast, \log \mQ  \rangle - \langle   1- \vx^\ast, \log (1 - \mQ)  \rangle,
			%&- \textstyle \sum_{i \in V_1, j \in V_2}  [\vx]_{ij}^\ast \log [\mQ]_{ij} - \\
			%& \sum_{i \in V_1, j \in V_2}  (1 - [\vx^\ast]_{ij}) \log( 1- [\mQ]_{ij} )
		\end{aligned}
	\end{equation}
	where $\langle : , : \rangle$ refers to the inner product operation for clarity. 
	
	At the testing stage, we deploy the Hungarian method \cite{Hungarian} to get 0-1 decision variable from the matrix $\mQ$. The Hungarian method is widely adopted for linear programming problems \cite{IPFP,pia,gmn,DGMC,DGFL}, with $\mathcal{O}(n^3)$ complexity, where $n$ is the node number of the input graph. Note that even though the Hungarian method consumes the highest complexity in our model, it takes less than 0.6 seconds when the input graphs are of size $n=1000$. We find that the real obstacle on the scalability of the model is the large memory footprint caused by the large matrix multiplication, see Section \ref{sec:dis2}.

\section{Ensemble model as a graph neural network}

\revised{In this section, we interpret our ensemble quadratic assignment model as a graph neural network (GNN) based on the observation in Sec.\ref{sec:QAP-Solver as a single-channel GNN} that the QAP-solver can be considered as a single-channel GNN on an association graph.}
We first connect the ensemble model to the multi-channel GNN and discuss the difference from the conventional GNN models. 
Then, we propose an affinity update scheme to enhance the matching performance,
and then introduce a random sampling strategy to reduce both the computational complexity and the GPU memory footprint to make our model suitable for large-scale problems. 
\revised{Finally, we analyze the computational complexity.}

\subsection{Ensemble model is a multi-channel GNN}

In the ensemble model, multiple channels are updated in parallel as in \eqref{eq:block}.
This can be seen as a multi-channel extension of \eqref{eq:update-rule-overall}.

Different from the conventional GNN models, where the convolution kernel $\mM$ is a learnable matrix, we consider $\mM$ as another input variable. 
For the graph matching problem, $\mM$ encloses the node affinity between two base graphs, which heavily depends on the input information and varies instance by instance.  
Therefore, we design the specialized GNN architecture  shown in Figure~\ref{overview}. 
The kernel $\mM$ is computed by the initialization module defined in \eqref{eq:initialization_module}, see Section \ref{sec:initialization_module}. 
All the ensemble blocks share the same kernel. 
Under this structure, given $\mM$, each channel of node features evolves independently. 
To promote information exchange among channels, we append a $1\times 1$ convolution layer, of which the $1\times 1$ kernel is a learnable parameter. 
By this design, our ensemble model inherits the high flexibility of GNNs and meanwhile utilizes the effectiveness of the classical graph matching algorithms.

\subsection{Affinity update scheme }
\label{sec:affinity-update}

In the ensemble model described above, the convolution kernel $\mM$ is shared among various modules, which limits the representational ability of the model to some extent.
Based on the GNN properties, we design an affinity update scheme to improve this situation.

Specifically, the pair-wise and node-wise affinity is updated according to
\begin{equation}\label{eq:ph1}\begin{aligned}
\mM^{(\ell)}_{i i^\prime, j j^\prime} &= \exp\Big[ - \vw^{(\ell)\T} \Big(\mV_{:, i, j}^{(\ell)} - \mV_{:, i^\prime, j^\prime}^{(\ell)} \Big)^2   \Big],
\\
\mM^{(\ell)}_{ii, jj} &= \exp \Big[ \vu^{(\ell)\T} \mV_{:, i, j}^{(\ell)} \Big],
\end{aligned}
\end{equation}
where $\vw^{(\ell)}$ and $\vu^{(\ell)}$ are two learnable parameter vectors, $\mV_{:, i, j}^{(\ell)}$ is the representation of the $ij$th association-node from the association graph, and the square in the first equation is an element-wise operation.
Intuitively, the two learnable parameters $\vw^{(\ell)}$ and $\vu^{(\ell)}$ controls the weight on how much different channels of the association-node features contributes to the pair-wise and node-wise affinity in the next layer. 
Thus, the inference rule in \eqref{eq:block} of the EnsembleBlock should be modified to 
\begin{align}  \label{eq:block_update} \nonumber
		&\bm{V}^{(\ell)}, \bm{M}^{(\ell)}   = \text{EnsembleBlock}^{(\ell)} \Big(  \bm{V}^{(\ell-1)} ~;~\bm{M}^{(\ell-1)} \Big),
\end{align}
where each block indexed by $\ell \in \{2, ..., L \}$ takes the feature tensor $\bm{V}^{(\ell-1)}$ and the updated affinity matrix $\mM^{(\ell-1)}$ from the previous ($\ell-1$th) block as input. 
When $\ell=1$, the input affinity matrix $\mM^{(0)}$ is the initial affinity matrix from the initialization module, see Section \ref{sec:initialization_module}.

\subsection{Scaling to large graphs via random sampling}
\label{sec:dis2}

	We present a random sampling strategy to make our model scalable to large graphs (those with as many as $>1000$ nodes).
	The general idea is to approximate the iteration equation \eqref{eq:iter-1} by only updating a random subset of items and directly copying the remaining entries from its corresponding input entries.
	This technique  not only lowers theoretical computational complexity but also reduces GPU memory consumption. 
	Moreover, the performance does not vary significantly.

	Specifically, we first generate a random mask $B^{(\ell)}_{c,i,j} \in \{0,1\}$ with $q^{(\ell)}_{i,j}$ probability being 1 for each layer $\ell \in \{1,...,L \}$, channel $c$, and position $i,j$.
	Note that the random mask $B$ is also not shared across different input data pairs.
	The probability is computed by
	$$q^{(\ell)}_{i, j} = [\bm{S}^{(\ell)}]_{i,  j}     / ( \sum_{k \tilde{k}} [\bm{S}^{(\ell)}]_{k \tilde{k}} ),$$
	where the sampling weight
	$[\mS^{(\ell)}]_{i, j} =\tfrac{1}{C}  \sum_{c=1}^{C} [\widetilde{\mV}^{(\ell-1)}]_{c,i,j}$ for $\ell > 1$, and $\widetilde{\mV}^{(\ell-1)}$ defined by \eqref{eq:matcher} is the internal representations in the previous block, \textit{e.g.},  the outputs of all solvers in the previous {ensemble block}. 
	For the first block, $\ell=1$, we set $[\mS^{(1)}]_{i, j} =[\mM]_{ii,jj}$, where $\mM$ is the affinity matrix generated from the initialization module.

	Then, for each channel $c$, we modify the QAP-solver so that it only updates the corresponding $[\widetilde{\vz}]_{i,j}$ in \eqref{eq:iter-1} if $B^{(\ell)}_{c,i,j}=1$, and copy the remaining entries from its input. Formally, we rewrite the update rule \eqref{eq:iter-1} to
	\begin{equation}
		\begin{aligned}
			\Big[\tilde{z}_{\ell+1} \Big]_{i,j} =  B^{(\ell)}_{c,i,j} \Big[ & \exp \Big( w_p \bm{M} \bm{z}_{\ell}
			+  w_z \log(\bm{z}_{\ell})
			\Big)\Big]_{i,j} \\
			&+ (1-B^{(\ell)}_{c,i,j}) [\widetilde{\vz}_{\ell}]_{i,j}.
		\end{aligned}
	\end{equation}

\minorrevised{The proposed sampling scheme is inspired by the neural network pruning \cite{NNP1,NNP2}, that the neurons with small-scale outputs and the connections with negligible weights are considered as uninformative, which can be pruned without significant impact on the performance. 
}

	\paragraph{\textbf{Approximated backward propagation}}
	The above random sampling strategy introduces a non-differentiable path along sampling weight matrix $\mS^{(\ell)}$ and the random mask $B^{(\ell)}_{c,i,j}$. 
	We solve this issue by proposing an approximated backward propagation.
	
	The weight matrix  $\mS^{(\ell)}$  affect the probability distribution of the random variable $B^{(\ell)}_{c,i,j}$ via $q^{(\ell)}_{i \tilde{i}}$, but  $B^{(\ell)}_{c,i,j}$ does not directly depends on $\mS^{(\ell)}$. There is no explicit backward gradient path along those two variables. It introduces a non-differentiable path along sampling weight matrix $\mS^{(\ell)}$ and the random mask $B^{(\ell)}_{c,i,j}$. We solve this issue by using the straight-through estimator (STE) to approximate the gradient \cite{STE}. Specifically, STE approximates the gradient $\frac{\partial {L}}{\partial \bm{S}^{(\ell)}_{i,j}} = \sum_{c,t,k}\frac{\partial {L}}{\partial  {B}^{(\ell)}_{c,t,k}} \frac{\partial  {B}^{(\ell)}_{c,t,k} }{\partial q^{(\ell)}_{t,k}} \frac{\partial q^{(\ell)}_{t,k} }{\partial \bm{S}^{(\ell)}_{i,j}}$ by assuming $\frac{\partial}{\partial q^{(\ell)}_{i \tilde{i}}} B^{(\ell)}_{c,i,j} \approx 1$. 
	
	Thus, we have 
	\begin{equation}
		\begin{aligned}
		\frac{\partial {L}}{\partial \bm{S}^{(\ell)}_{i,j}} &\approx \sum_{c,t,k}\frac{\partial {L}}{\partial  {B}^{(\ell)}_{c,t,k}} \frac{\partial q^{(\ell)}_{t,k} }{\partial q^{(\ell)}_{t,k}} \frac{\partial q^{(\ell)}_{t,k} }{\partial \bm{S}^{(\ell)}_{i,j}}
		\\
		%&= \sum_{c,t,k}\frac{\partial {L}}{\partial \bm{B}^{(\ell)}_{c,t,k}} \frac{\partial q^{(\ell)}_{t,k} }{\partial \bm{S}^{(\ell)}_{i,j}}
		%\\
		&= \frac{1}{\sum \bm{S}^{(\ell)}} \sum_{c,t,k}\frac{\partial {L}}{\partial  {B}^{(\ell)}_{c,t,k}} (\delta_{t:i}\delta_{k:j} - q^{(\ell)}_{i,j}),
			%\Big[\tilde{z}_{\ell+1} \Big]_{i,j} =  B^{(\ell)}_{c,i,j} \Big[ & \exp \Big( w_p \bm{M} \bm{z}_{\ell} +  w_z \log(\bm{z}_{\ell})  \Big)\Big]_{i,j} \\
			%&+ (1-B^{(\ell)}_{c,i,j}) [\widetilde{\vz}_{\ell}]_{i,j}.
		\end{aligned}
	\end{equation}
	where $\delta_{t:i}$ is the Dirac delta function that $\delta = 1$  when $t$ equals $i$, and $\delta = 0$ otherwise. 

 \minorrevised{The straight-through estimator (STE) \cite{STE} used above estimates the gradient via network quantization and network pruning. It considered  back-propagating gradients through a piece-wise constant function that has zero gradients almost everywhere. The STE  ignores the quantization (discretization) function in the backward pass and passes the gradient through as if it acts as an identity function. Yin et. al. \cite{Yin} provides a theoretical justification  of STE, which proved that the expected coarse gradient given by the STE-modified chain rule converges to a critical point of the population loss minimization problem. }

	%\subsection{Model summary}
	%The time complexity of the vanilla version of our approach is $\mathcal{O}(k^2n^2)$, which is proportional to the number of hyper-edges. The time cost of ours is slightly higher than the cost of the previous SOTA model - DGMC, which is $\mathcal{O}(N^2+2kN)$ \cite{DGMC}.

\subsection{\revised{Complexity analysis}}

The update rule \eqref{eq:iter-1}, the most costly part of the overall complexity of the model, has the complexity  $\mathcal{O}(K^2n^2)$, where $n$ is the  size of the node set of the input graphs $G_1$ and $G_2$, and $K$ is the largest node degree in $G_1$ and $G_2$.
	The dominating part is the matrix-vector product $ \bm{M} \bm{z}_t $ in \eqref{eq:iter-1}.
	In practice, we set the number of sampled entries in $\bm{z}$ to be updated proportionally to  $n\sqrt{n}$ in each QAP-solver, which is much less than the number of all entries $n^2$ in $\bm{z}$. Thus, the complexity \minorrevised{of the update \eqref{eq:iter-1} }  is reduced  to $\mathcal{O}(K^2 n \sqrt{n})$.
	In the view of graph neural network, this sampling operation is equivalent to  using the message-passing mechanism to update the information of a few randomly selected association-nodes from the association graph. \revised{In Section 6.5.8, we further investigate practical running time at inference stage.}

	\minorrevised{ In addition, we note that the overall theoretical time complexity of the whole EQAN model is still $\mathcal{O}(n^2)$,  which is dominated theoretically by the Sinkhorn normalization step and the final decision layer.  On the other hand, the sampling scheme, as analyzed above, reduces the number of nodes to be updated from $n^2$ to $n\sqrt{n}$. The major target of this sampling strategy is to reduce the GPU memory footprint, which makes the model scales up to   thousand of nodes in a single 12GB GPU card.}
 
 %this is because there are a few operations are still with the complexity proportional to $\mathcal{O}(n^2)$, such as the Sinkhorn normalization step and the final decision layer. Therefore, we can see that the overall operating speed has not been significantly improved as shown in Fig \ref{speed}. Fortunately, thanks to the sampling strategy, the model can have less memory footprint and be extended to larger graphs, also check Fig \ref{speed} for details. 

\section{Experiments}
 
	\begin{figure*}[htp]
		\centering
		\includegraphics[width=1\linewidth]{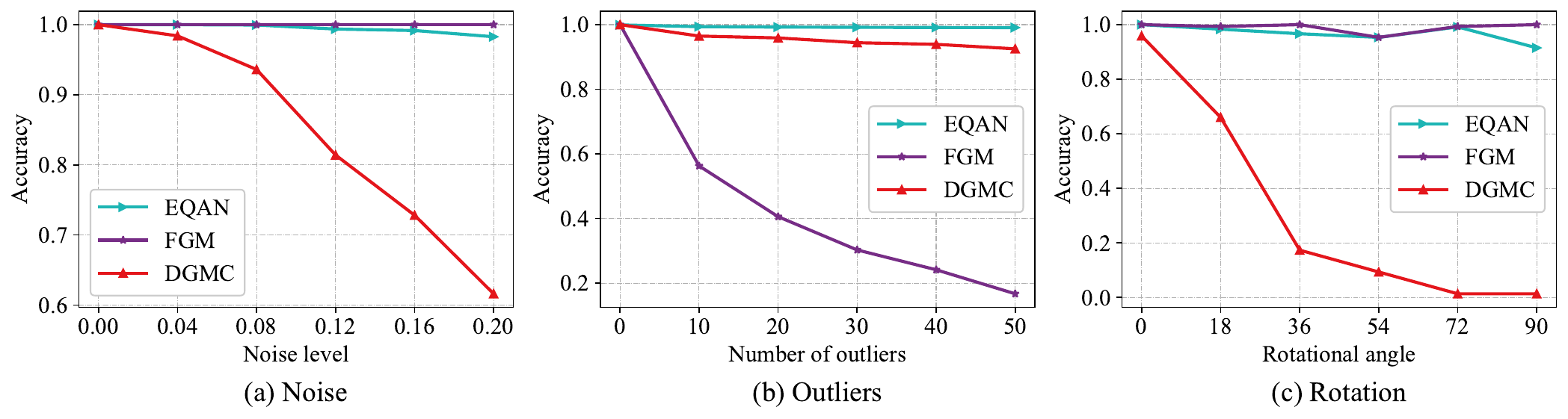}
		\caption{Robustness to noise, outliers, and random rotation. (a) The noise level $\sigma$ changes from 0.0 to 0.2, with $n_\text{in} = 50$, and $n_\text{out} = 0$.
		(b) We add a number of outliers $n_\text{out}$ from $0$ to $50$ with $n_\text{in} = 35$, and $\sigma = 0.1$.
		(c) The maximum rotational angle ranges from $0$ to $90$ with $n_\text{in} = 30$, $n_\text{out} = 15$, and $\sigma = 0.1$. }
		\label{fig:robustness}
	\end{figure*}

\begin{figure}[htp]
	\centering
	\includegraphics[width=1\linewidth]{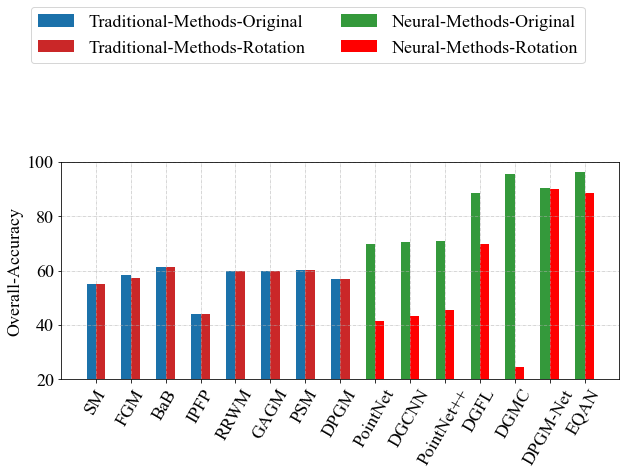}
	\caption{\label{fig:unrobust deepGM} 
 \revised{ Robustness of matching accuracy to random rotations. Left/right bar shows the accuracy on the original/rotated data respectively in each method.  Traditional
solvers is robust to global rotation but less accurate, whereas
existing GNN methods improves the accuracy noticeably
but they are very sensitive to global rotations. Our methods (DPGM-net and EQAN) inherit the advantages of both sides.}
 %Comparison of robustness among different graph matching approaches: PointNet \cite{Pointnet}, DGCNN \cite{DGCNN}, PointNet++ \cite{Pointnet++}, DCP \cite{DCP}, DGFL \cite{DGFL}, DGMC \cite{DGMC}. 
 }
	%\vspace{-0.6cm}
\end{figure} 

	We conducted a series of experiments to evaluate the performance of the proposed method.
    	First, we test its robustness in three aspects: raw feature noise, outliers, and random rotation. 
	Then, we compare the performance with existing methods in three tasks: geometric graph matching task, semantic feature matching, and few-shot 3D shape classification. 
	Finally, we present ablation studies to investigate the effects of internal settings of our approach.
	All experiments were implemented in PyTorch \cite{pytorch} and run on a computer with the Intel Xeon E5-2680 processor and the NVIDIA TITAN GPU. 
	
	We use the proximal QAP-solver (DPGM) to construct the ensemble network.
	We implement three variants of our ensemble model, namely the naive ensemble quadratic assignment network  (EQAN), the network with the affinity update scheme (EQAN-U) in Section \ref{sec:affinity-update}, and the network with the random sampling strategy (EQAN-R) in Section \ref{sec:dis2} respectively.
	The reported results of our methods in experiments are the average score of 5 independent runs  (20 runs for few-shot 3D shape classification).

	\paragraph{Training with  synthetic data} 
	In the robustness test, the 2D geometric graph matching task, and the 3D-shape few-shot  classification task, we train the model solely on the synthetic data \cite{DGFL}. 
	This setting is more challenging to test generalization ability, and it is close to the practical situation when node-level annotated data is expensive or not available for training. 
	In those experiments, we generate the reference graph ${G}_{1}$ with $n_\text{in}$ inliers nodes embedded in a d-dimensional Euclidean space ($d=2,3$).
	The node feature information is uniformly drawn from $[-1, 1]^d$ and each node is connected to its top-5 nearest neighbors. 
	Then, the query graph $G_2$ is generated by adding Gaussian noise to the nodes of a duplicated graph with zero mean and standard variance $\sigma$. In addition, we introduce $n_{\text{out}}$ randomly uniform outlier nodes to $G_2$.

    \paragraph{Evaluation metric}
	For geometric graph matching and semantic feature matching, we evaluate the matching accuracy by calculating the ratio of the number of correctly matched node pairs to the total number of ground truth node pairs \cite{pia}.
	For the few-shot shape classification, the classification accuracy is computed by calculating the ratio of the number of correctly classified query samples to the query set size \cite{few-shot-Neurocomputing,few-shot-Berkeley}.

	\subsection{Robustness test}

	\label{sec:robustness}

	We test the sensitivity of our model to the feature noise, a number of outliers, and 2D rotations in the 2D graph matching setting, where the data generation settings of $\sigma, n_\text{in}, n_\text{out}$ are listed in the caption of Figure \ref{fig:robustness}.
	Experimental results are shown in Figure \ref{fig:robustness}. We compare our method with the classical algorithm, FGM \cite{fgm}, and the GNN-based approach, DGMC  \cite{DGMC}.
	The affinity constructing of FGM is following the settings in \cite{fgm}.
	The settings for DGMC are the same as the geometric graph matching settings in \cite{DGMC}.  
    	For our model, the block number in our model is $L=5$, and each block consists of $C = 32$ proximal QAP-solvers \cite{AAAI2021}.
	In training, we use the Adam optimizer \cite{Adam} with a learning rate of 1e-4, batch size of 8, total iterations of 80000, and a warm-up stage of 500 iterations with a learning rate of 1e-10.

	In the first test of noise robustness, the classical algorithm FGM is stable against  the noise and our approach's performance drops slightly  when increasing the noise, whereas the GNN-based method, DGMC is very sensitive to noise.
	In the second test of outlier robustness, the classical algorithm FGM deteriorates remarkably with outliers, while our approach and the DGMC work stably.
	In the third test of rotation sensitivity, the traditional algorithm FGM and our approach are almost unaffected to the rotation but the performance of DGMC drops remarkably.

 \revised{ In addition, we compare with other 14 peer methods including both types of traditional optimizer and GNN-based approaches in term of average matching accuracy under random rotations between $0$ to $90$ degrees. Results are shown in Figure~\ref{fig:unrobust deepGM}. Similar experiments on Willow-Object dataset were reported in our conference work \cite{AAAI2021}.  In general, traditional solvers is robust to global rotation but less accurate, whereas previous GNN methods improve the accuracy noticeably but they are very sensitive to global rotations.
}

	The above results verify that our approach not only inherits  the strong robustness of traditional algorithms but also preserves the  high prediction accuracy of graph neural network methods.

%\begin{figure*}[tp]
%		\centering
		%\includegraphics[width=1\linewidth]{chart/chart.pdf}
%		\includegraphics[width=0.68\linewidth]{fig/cmu-test.eps}

%		\caption{ \label{cmu} Geometric graph matching results on CMU-House dataset. According to the number of outliers, the experiment is divided into two parts: outliers = 0 and outliers = 10. }
		%\vspace{-0.6cm} %basic-layer.pdf
%	\end{figure*}

	\begin{table*}[tp]
\caption{\label{tb:pf} Geometric graph matching accuracy on PASCAL-PF \cite{Willow}.  All data-driven matching approaches are trained on 2D synthetic random graphs and then directly applied to the dataset. The results of Pointnet \cite{Pointnet} are reported in \cite{DGFL}.}
\scriptsize %\vspace{-1.0cm}
\begin{center} \resizebox{\textwidth}{!}{\begin{tabular}{l|ccccccccccccccccccccc}
\toprule[2pt]
\textbf{PASCAL-PF} & aero &bike &bird &boat &bottle &bus &car &cat &chair &cow &table &dog &horse &mbike &person &plant &sheep &sofa &train &tv &mean \\
\bottomrule[0.1pt]
\toprule[0.1pt]
${ \text{PointNet}}$ \cite{Pointnet}  &54.8 &70.2 &65.2 &73.7 &85.3 &90.0 &73.4 &63.4 &55.2 &78.4 &78.4 &52.5 &58.0 &64.2 &57.4 &68.9 &50.5 &74.0 &88.1 &91.9 &69.7  \\
${\text{DGFL}}$ \cite{DGFL}   & 76.1 & 89.8  &93.4  &96.4 &96.2 &97.1 &94.6 &82.8 &89.3 &96.7 &89.7 &79.5 &82.6  &83.5 &72.8 &76.7 &77.1 &97.3 &98.2 &99.5  &{88.5} \\
${\text{DGMC}}$ \cite{DGMC}  &81.1 &92.0 &94.7 &100.0 &99.3 &99.3 &98.9 &97.3 &99.4 &93.4 &100.0 &99.1 &86.3 &86.2 &87.7 &100.0 &100.0 &100.0 &100.0 &99.3 &95.7         \\
$\text{DPGM-Net}$ \cite{AAAI2021} &79.3 &92.5 &97.7 &96.4 &94.6 &96.4 &96.3 &95.7 &98.3 &95.0 &76.3 &90.7 &85.1 &84.7 &88.8 &65.2 &100.0 &100.0 &100.0 &67.2 &90.6 \\
\midrule[0.5pt]
$\text{without-ensemble}$    &48.4 &74.0 &69.3 &49.6 &46.1 &58.3 &73.2 &49.1 &62.5 &57.2 &41.9 &49.0 &61.3 &64.4 &42.4 &35.7 &53.3 &59.8 &63.2 &37.5 &57.0 \\
$\text{EQAN-R}$ &85.7 &90.6 &86.9 &98.3 &90.1 &88.0 &98.6 &94.5 &91.7 &91.8 &99.3 &87.7 &79.1 &79.8 &83.1 &88.1 &95.8 &93.5 &85.1 &93.2 &90.3 $\pm 0.8$ \\
$\text{EQAN}$  &87.2 &91.5 &94.6 &98.4 &99.4 &99.1 &98.9 &99.5 &99.6 &95.7 &100.0 &99.6 &90.5&89.4 &88.7 &100.0 &100.0 &99.2 &99.6 &99.6 &\textbf{96.2} $\pm 0.3$ \\%{$(\uparrow 5.6)$} \\
$\text{EQAN-U}$ & 87.7 & 91.1 & 90.8 & 98.8 & 99.8 & 98.9 & 99.1 & 97.0 & 100.0 & 96.0 &100.0 &100.0 & 90.8 & 89.6 &89.1 &100.0 &100.0 & 99.6 & 99.1 &100.0 &\textbf{96.4} $\pm 0.4$\\
\bottomrule[2pt]
\end{tabular}}
\end{center}
\end{table*}

\begin{table*}[http]
\caption{\label{tb:pascal} Semantic feature matching accuracy on PASCAL VOC Keypoints. Results of GMN, PCA, and CIE are taken from \cite{CIE}. The results of DGMC, NGM+, GLMNet are taken from \cite{blackbox}. The BBGM is implemented based on the official repository in \cite{blackbox}. The results of SIGMA are taken from Table 5 in the original article \cite{SIGMA}.
\revised{The per-class results of GAMnet \cite{GAMnet} and ASAR-GM \cite{ASAR-GM} are not available in their original papers.}
}
\scriptsize
\begin{center}
\resizebox{\textwidth}{!}{\begin{tabular}{l|cccccccccccccccccccc|c}
\toprule[2pt]
method & aero &bike &bird &boat &bottle &bus &car &cat &chair &cow &table &dog &horse &mbike &person &plant &sheep &sofa &train &tv &mean \\
\bottomrule[0.1pt]
\toprule[0.1pt]
GMN \cite{gmn}    &31.9  &47.2  &51.9  &40.8  &68.7  &72.2  &53.6  &52.8  &34.6  &48.6  &72.3  &47.7  &54.8  &51.0  &38.6  &75.1  &49.5  &45.0  &83.0  & 86.3  & 55.3  \\
PCA  \cite{pia}  &40.9  &55.0 & 65.8  &47.9  &76.9  &77.9  &63.5  &67.4  &33.7  &65.5  &63.6  &61.3  &68.9  &62.8  &44.9 & 77.5  &67.4  &57.5  &86.7  &90.9  &63.8          \\
CLDGM  \cite{CLDGM}  &51.0 &64.9 &68.4 &60.5 &80.2 &74.7 &71.0 &73.5 &42.2 &68.5 &48.9 &69.3 &67.6 &64.8 &48.6 &84.2 &69.8 &62.0 &79.3 &89.3 &66.9        \\
LCSGM  \cite{LCSGM}  &46.9 &58.0 &63.6 &69.9 &87.8 &79.8 &71.8 &60.3 &44.8 &64.3 &79.4 &57.5 &64.4 &57.6 &52.4 &96.1 &62.9 &65.8 &94.4 &92.0 &68.5    \\
NGM+ \cite{NGM}  &50.8 &64.5 &59.5 &57.6 &79.4 &76.9 &74.4 &69.9 &41.5 &62.3 &68.5 &62.2 &62.4 &64.7 &47.8 &78.7 &66.0 &63.3 &81.4 &89.6 &66.1 \\
GLMNet \cite{GLMNet}  &52.0 &67.3 &63.2 &57.4 &80.3 &74.6 &70.0 &72.6 &38.9 &66.3 &77.3 &65.7 &67.9 &64.2 &44.8 &86.3 &69.0 &61.9 &79.3 &91.3 &67.5 \\
CIE \cite{CIE}  &51.2 &69.2 &70.1 &55.0 &82.8 &72.8 &69.0 &74.2 &39.6 &68.8 &71.8 &70.0 &71.8 &66.8 &44.8 &85.2 &69.9 &65.4 &85.2 &92.4 &68.9\\
IA-DGM \cite{IA-GM} &53.9  &67.7  &68.8  &60.2 &80.3 &75.1  &76.9  &72.0  &40.2  &65.6  &79.6  &65.4  &66.2  &67.4   &46.3 &87.4 &65.4  &58.5 &89.4 &90.7 &68.8 \\
QC-DGM \cite{QCDGM}  &49.6 &64.6 &67.1 &62.4 &82.1 &79.9 &74.8 &73.5 &43.0 &68.4 &66.5 &67.2 &71.4 &70.1 &48.6 &92.4 &69.2 &70.9 &90.9 &92.0 &70.3\\
DGMC \cite{DGMC}  &50.4 &67.6 &70.7 &70.5 &87.2 &85.2 &82.5 &74.3 &46.2 &69.4 &69.9 &73.9 &73.8 &65.4 &51.6 &98.0 &73.2 &69.6 &94.3 &89.6 &73.2 \\
BBGM \cite{blackbox} &61.5 &75.0 &78.1 &80.0 &87.4 &93.0 &89.1 &80.2 &58.1 &77.6 &76.5 &79.3 &78.6 &78.8 &66.7 &97.4 &76.4 &77.5 &97.7 &94.4 & 80.0  \\
SIGMA \cite{SIGMA} &55.1  &70.6 &57.8 &71.3 &88.0 &88.6 &88.2 &75.5 &46.8 &70.9 &90.4 &66.5 &78.0 &67.5 &65.0 &96.7 &68.5 &97.9 &94.3 &86.1 &76.2 \\
\revised{DLGM} \cite{DLGM} &64.7  &78.1  &78.4  &81.0 &87.2 &94.6 &89.7 &82.5   &68.5   &83.0 &93.9 &82.3   &82.8 &82.7 &69.6 &98.6 &78.9   &88.9 &97.4 &96.7 &\textbf{83.8} \\
\revised{NGMv} \cite{NGM} &59.9 &71.5 &77.2 &79.0 &87.7 &94.6 &89.0 &81.8 &60.0 &81.3 &87.0 &78.1 &76.5 &77.5 &64.4 &98.7 &77.8 &75.4 &97.9 &92.8 &80.4 \\
\revised{GAMnet} \cite{GAMnet} &-  &- &- &- &- &- &- &- &- &- &- &- &- &- &- &- &- &- &- &- &80.7 \\
\revised{ASAR-GM} \cite{ASAR-GM} &-  &- &- &- &- &- &- &- &- &- &- &- &- &- &- &- &- &- &- &- &\textbf{81.2} \\
\midrule[1pt]
${\text{without-ensemble}}$ &49.7 &64.8 &57.3 &{57.1} &{78.4} &{81.2} &{66.3} &{69.6} &{45.2} &63.4 &{64.2} &{63.5} &67.9 &62.0 &{46.8} &{85.5} &{70.7} &{58.1} &{88.9} &{89.9} &{66.5 $\pm 0.5$}         \\
EQAN-R &  59.9  &  59.3  &  71.2  &  63.5  &  85.9  &  77.6  &  71.1  &  83.2  &  46.1  &  58.5 & 99.2 & 72.9 & 80.8 & 65.4 & 58.6 & 79.5  &  74.2  &  51.0  &  90.4  &  80.9  &  71.5 $\pm 0.9$\\
${\text{EQAN}}$ &66.4 &74.4 &82.6 &71.0 &88.7 &87.7 &86.4 &80.4 &53.4 &66.6 &100.0 &80.3 &78.0 &76.7 &64.6 &96.5 &83.7 &62.2 &100.0 &95.5 &\textbf{79.8 $\pm 0.5$}  \\
{${\text{EQAN-U}}$}   &68.5  &76.9  &81.9  &71.2  &89.1  &91.3  &82.0  &82.9  &52.0  &70.9   &100.0  &80.9  &78.8  &79.3  &66.7  &95.7  &87.4  &65.7  &100.0  &96.8  &\textbf{80.9 $\pm 0.4$}  \\
\bottomrule[2pt]
\end{tabular}}
\end{center}
\end{table*}

\subsection{Geometric graph matching}

	We compare the performance of geometric graph matching on the PASCAL-PF \cite{Willow}, which consists of 1351 image pairs within 20 classes. 
	Each image pair contains 4-17 manually labeled ground-truth correspondences. 
	This task aims to estimate the correspondence with only the coordinates of keypoints \cite{DGFL,DGMC,fgm,RRWM,KerGM}. 
	The experimental results are reported  in Table \ref{tb:pf}.

	We implemented the recently proposed  data-driven approaches as baselines, including the DGFL \cite{DGFL}, DGMC \cite{DGMC}, Pointnet \cite{Pointnet}, and the DPGM-Net \cite{AAAI2021} which is an end-to-end network that utilizes the proximal solver with a graph neural network \cite{DGFL} to generate features for affinity matrix computing.

\begin{figure*}[tp]
		\centering
		\subfigure[Ours] {
			\includegraphics[width=0.29\linewidth]{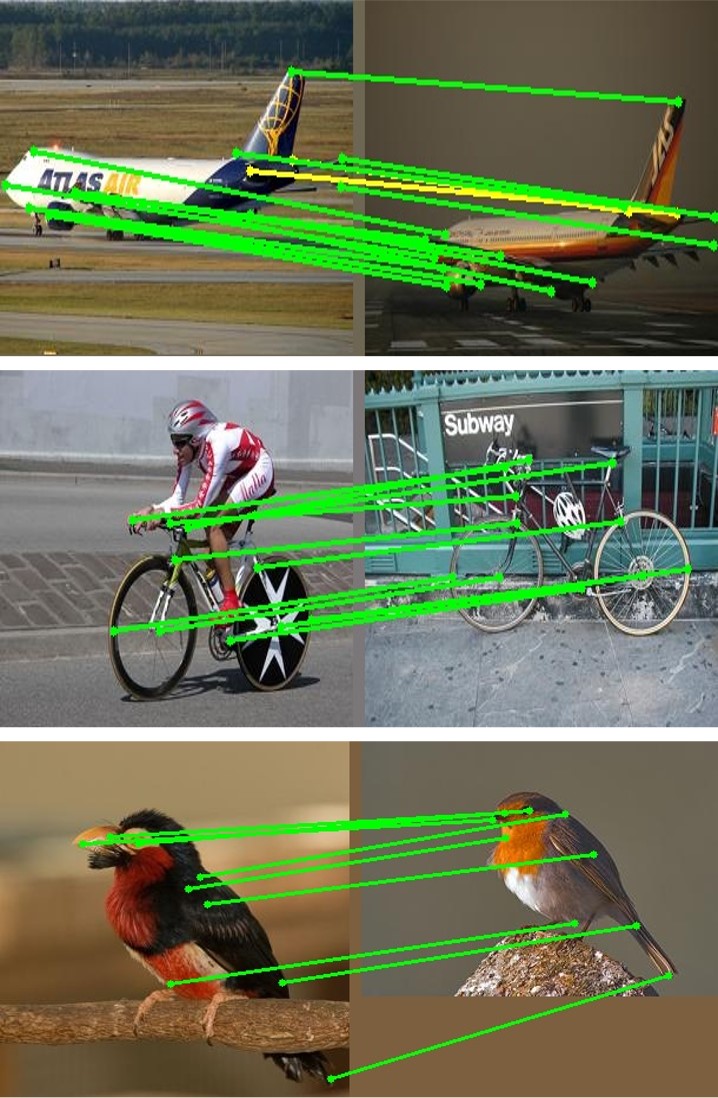}
		}
		\hspace{-0.2cm}
		\subfigure[DGMC \cite{DGMC}] {
			\includegraphics[width=0.29\linewidth]{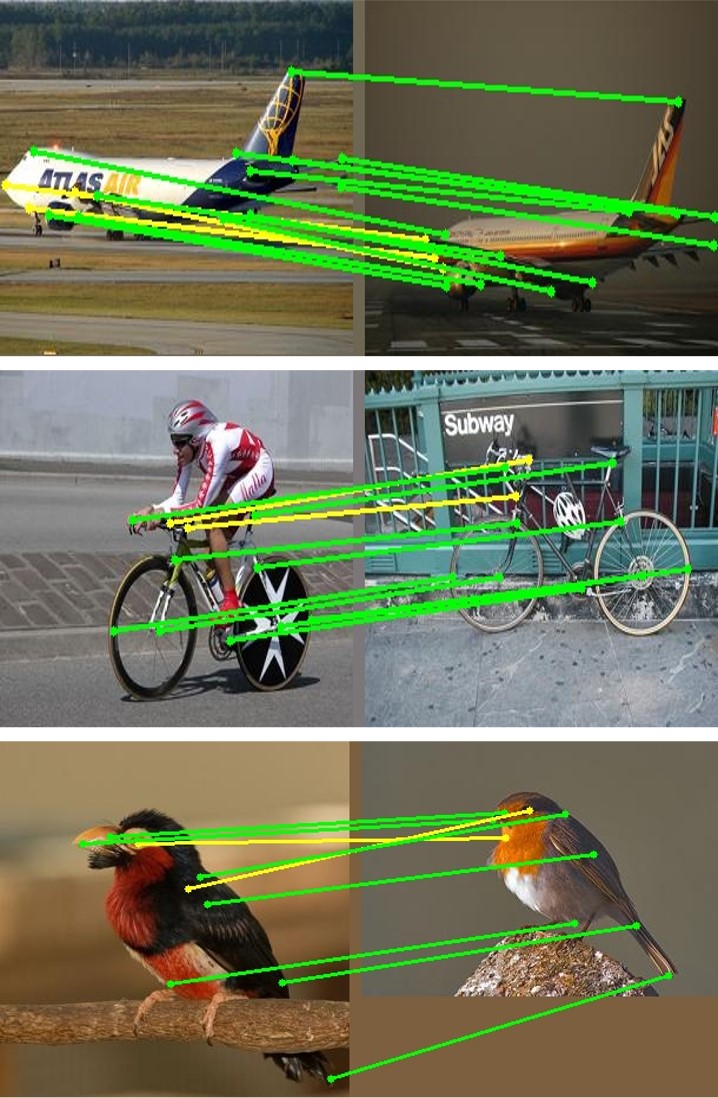}
		}
		\hspace{-0.2cm}
		\subfigure[FGM \cite{fgm}] {
			\includegraphics[width=0.29\linewidth]{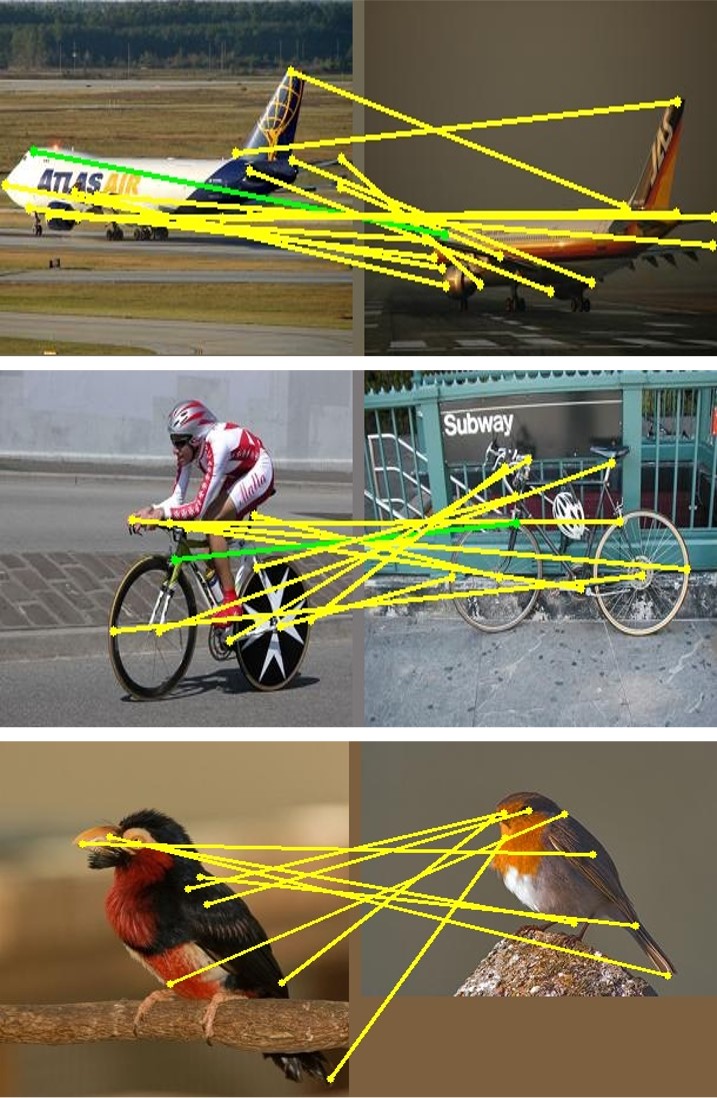}
		}
		\caption{Typical matching results on PASCAL-PF. The green line indicates the matching is correct, and yellow means wrong. FGM \cite{fgm} is one of the classical traditional graph matching methods. DGMC \cite{DGMC} and ours are both trained on synthetic graphs only.}
\label{img_pf}
\end{figure*}

	In geometric graph matching, our model directly uses the normalized coordinates as the input features $\bm{F}_1$ and $\bm{F}_2$ for the initialization module.
	The specific model structure is consistent with the structure used in the robustness test.
	All the data-driven baselines were trained on 2D synthetic random graphs with $\sigma = 0.005, n_\text{in} = 50, n_\text{out} = 15$, and then directly applied to the real-world dataset. The optimizer settings in training are identical to the ones used in the robustness test experiment.
	In Table \ref{tb:pf}, ``without-ensemble" denotes the model that the single QAP-solver is applied as a stand-alone traditional solver without any GNN feature extractor. It performs roughly 40 percent worse than the proposed ensemble model.
	Our ensemble model achieves the best performance, outperforming the previous best geometric matching method, DGMC \cite{DGMC}. Some examples of matching results on PASCAL-PF are shown in Figure \ref{img_pf}.

 \revised{Additionally, we conducted an experiment on the CMU-House dataset. 
For the traditional matching algorithms including PSM \cite{PSM}, GAGM \cite{GAGM}, RRWm \cite{RRWM}, SM \cite{sp1}, we build graphs by using these landmarks as nodes and adopting Delaunay tri-angulation to generate edges. 
For the deep learning matching approaches including DGFL \cite{DGFL}, DGMC \cite{DGMC}, PointNet \cite{Pointnet}, and our EQAN, we normalize the keypoint coordinates by subtracting the statistic mean and dividing by its statistic variance. 
We report the final average matching accuracy in Figure 3 under the frame gap setting of 10:10:100 frames. 
The results show that our method can keep a relatively good performance even when the number of inliers is small, and the frame gap is large.
}
\begin{figure}[thtp]
	\centering
	\includegraphics[width=1\linewidth]{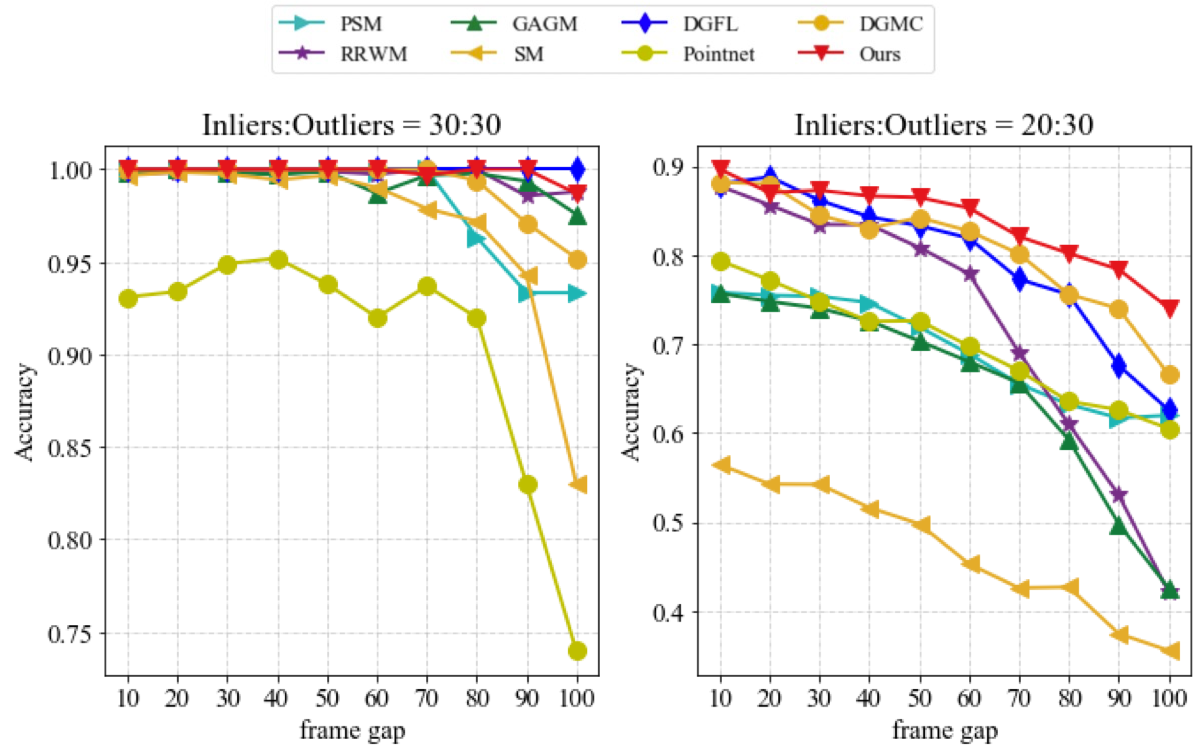}
	\caption{\label{fig:cmu eqan} \revised{Comparison of matching accuracy on the CMU-House dataset.}}
	%\vspace{-0.6cm}
\end{figure}

\begin{table*}[tp]
\caption{\label{tb:spair} Semantic feature matching accuracy on Spair-71K. Results of DGMC, BBGM are taken from \cite{blackbox}. For SIGMA \cite{SIGMA}, the mean accuracy is $79.8 \pm 0.2$, while the accuracy score per-class is not provided in \cite{SIGMA}. }
\scriptsize
\begin{center}
\resizebox{\textwidth}{!}{\begin{tabular}{l|cccccccccccccccccc|ccc}
\toprule[2pt]
method & aero &bike &bird &boat &bottle &bus &car &cat &chair &cow  &dog &horse &mbike &person &plant &sheep  &train &tv &mean \\
\bottomrule[0.1pt]
\toprule[0.1pt]
DGMC \cite{DGMC}   &54.8&44.8&80.3&70.9&65.5&90.1&78.5&66.7&66.4&73.2&66.2&66.5&65.7&59.1&98.7&68.5&84.9&98.0&72.2  \\
BBGM \cite{blackbox}   &66.9&57.7&85.8&78.5&66.9&95.4&86.1&74.6&68.3&78.9&73.0&67.5&79.3&73.0&99.1&74.8&95.0&98.6&78.9  \\
\revised{SIGMA} \cite{SIGMA} &-      &-    &-    &-    &-  &-  &-  &-  &-  &-  &-  &-  &-  &-  &-  &- &-  &-  &79.8\\
\revised{DLGM} \cite{DLGM} &70.4 &66.8 &86.7 &81.7 &69.2 &96.4  &85.8 &79.5 &78.4 &84.0 &79.4 &69.4 &84.5 &76.6   &99.1 &75.9 &96.4 &98.5 &\textbf{82.0} \\
\midrule[1pt]
without-ensemble    &31.7      &72.3    &76.0    &42.7    &54.3  &76.9  &93.3  &60.0  &53.0  &62.2  &50.2  &40.7  &55.3  &52.2  &87.2  &70.9 &99.5  &82.9  &65.6  $\pm 0.1$ \\
EQAN-R  &51.9  &55.6  &78.7  &72.3  &59.0  &89.2  &78.5  &75.3  &52.2  &73.1  &71.0  &69.8  &80.8  &54.7  &93.7  &66.5  &80.4  &89.0  &71.8 $\pm 0.8$  \\
EQAN &65.3 &62.5 &86.7 &83.5 &69.7 &95.5 &92.3 &77.5 &66.5 &78.2 &75.8 &73.4 &81.3 &64.6 &99.3 &74.5 &93.4 &99.3 &{79.9 $\pm 0.5$}  \\
EQAN-U  &65.5  &65.1  &81.1  &84.2  &71.7  &97.9  &93.4  &78.1  &67.7  &80.1  &72.8  &74.3  &83.2  &67.3  &100.0  &75.7  &93.9  &100.0  &{80.7  $\pm 0.3$}\\
\bottomrule[2pt]
\end{tabular}}
\end{center}
\end{table*}

\subsection{Semantic feature matching}

	We evaluate the performance of semantic feature matching on PASCAL-VOC and Spair-71K datasets following the experimental protocol in \cite{gmn,pia,blackbox,DGMC,NGM}.

	\subsubsection{PASCAL-VOC Keypoint}
	This dataset \cite{pascal} contains 20 classes of instances and 7,020 annotated images for training and 1,682 for testing. Berkeley's annotations provide labeled keypoint locations \cite{pascal}.
	Following \cite{gmn,RPCA,pia}, we crop the bounding box around the object in the image and resized the cropped box to $256\times 256$.
	For baselines, we choose the following methods: PCA \cite{pia}, GMN \cite{gmn}, CIE \cite{CIE}, CLDGM \cite{CLDGM}, LCSGM \cite{LCSGM}, DGMC \cite{DGMC}, BBGM \cite{blackbox}, NGM+ \cite{NGM}, QC-DGM \cite{QCDGM}, IA-DGM \cite{IA-GM}, GLMNet \cite{GLMNet}, SIGMA \cite{SIGMA} and DLGM \cite{DLGM}.

    \paragraph{Training protocal} By following \cite{gmn,pia,blackbox,DGMC,NGM}, we build the graph by setting the keypoints as nodes and then using Delaunay triangulation to connect landmarks.
    \revised{The node features are generated from the conv4-2 and the conv5-1 layer of a pre-trained VGG-16 \cite{vgg}, which is fixed without further fine-tuning during training of GNN.} Since the width and height of the feature map are smaller than the given image, we use the bilinear interpolation to approximately recover the node feature of the given landmark pixel in the image.
    
    For our model, by following \cite{blackbox,DGMC,DLGM,SIGMA}, we adopt a SplineCNN network \cite{SplineCNN} as the GNN module in the initialization module to further refine the input semantic features from the VGG feature extractor. For constructing the ensemble assignment module, we set the number of proximal QAP-solver in each block to be $C = 32$ and set the block number $L = 5$.
    We also adopt the single proximal QAP-solver (DPGM) as the differentiable solver attached after the VGG-16 feature extractor as a baseline, see the item ``without-ensemble''. It is  also trained on the same dataset in an end-to-end fashion.

\begin{figure*}[tp]
		\centering
		\includegraphics[width=0.91\linewidth]{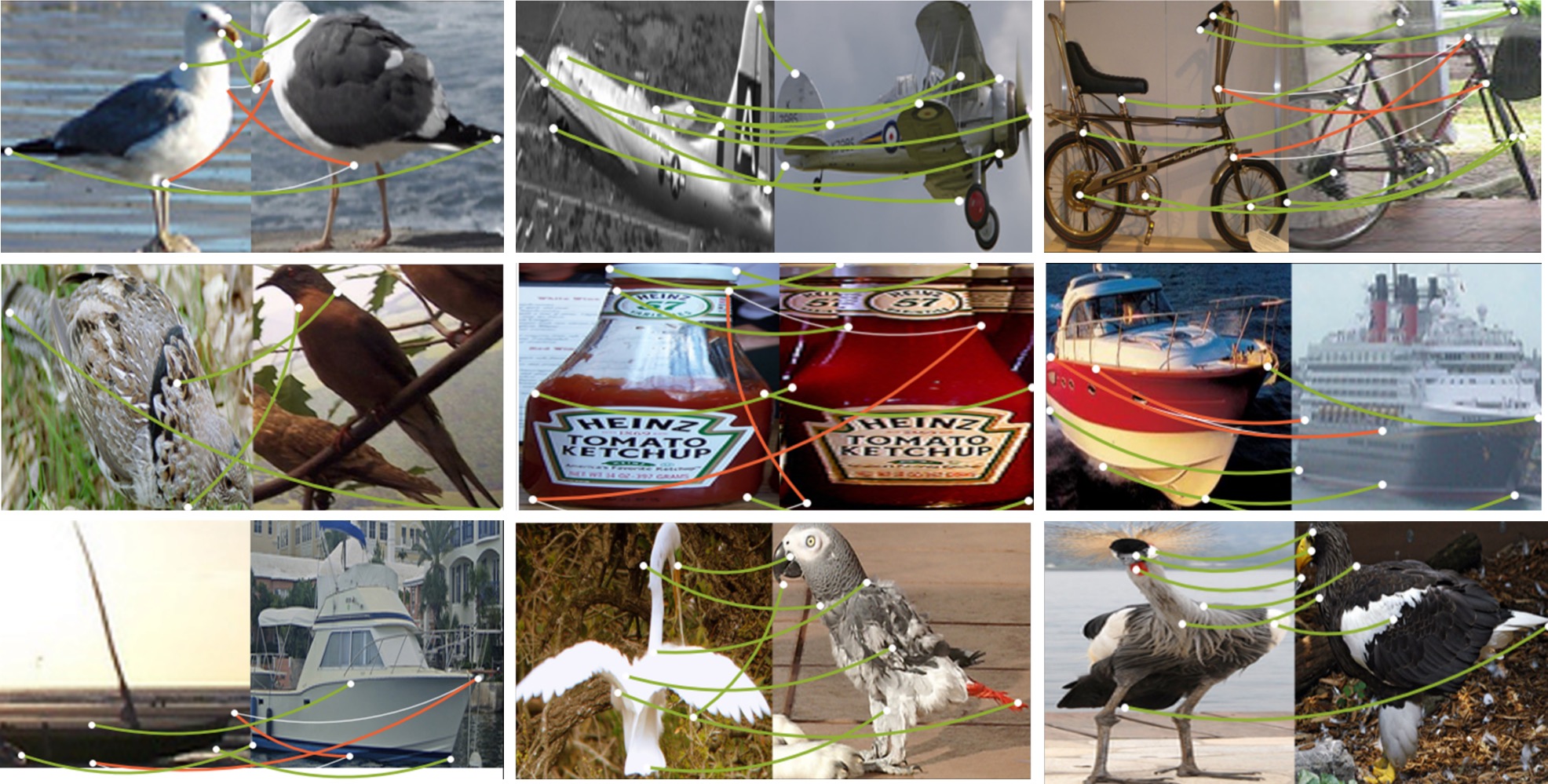}
		\caption{Example semantic feature matching results of the proposed approach on PASCAL-VOC. Our method can handle pairs of images seen from very different viewpoints. The green line indicates the matching is correct, and red means wrong. The visualization code is based on the official code repository of \cite{blackbox}. }
		\label{img_voc}
		%\vspace{-0.6cm} %basic-layer.pdf
	\end{figure*}

    Table \ref{tb:pascal} gives the experimental results.
    Compared to the ``without-ensemble'', our ensemble model with multiple proximal solvers (EQAN, EQAN-U) exhibits a considerable performance improvement of about 13\%.
    The EQAN-U model outperforms DGMC \cite{DGMC}, BBGM method \cite{blackbox}, and SIGMA \cite{SIGMA}, by 7.7\%, 0.9\%, and 4.7\% in matching accuracy respectively.  
    Our EQAN achieves comparable performance with NGMv2 \cite{NGM}, GAMnet \cite{GAMnet} and ASAR-GM \cite{ASAR-GM}. Visual examples of the matching results on PASCAL-VOC are shown in Figure \ref{img_voc}.
    \minorrevised{Our work, mainly focused on the network architecture 
    based on the connections between the ensemble of classical algorithms and  the multi-channel graph neural networks, has the potential of further improvement by combing the idea of adversarial training used in ASAR-GM, but it  requires  more carefully-designed architecture with the corresponding training method. We leave this study for future work. 
    }

    \paragraph{EQAN v.s. DLGM}  DLGM \cite{DLGM} proposed a complex but novel graph feature extractor, the core of which is to use a neural network to learn potential graph structure. It achieved an amazing accuracy of 83.8\% on Pascal-VOC Keypoint and 82.0\% on Spair-71K. However, the code of DLGM is still not open to the public. 
    Considering the solver module in DLGM is just a simple traditional quadratic assignment algorithm, we believe that our deep ensemble model can also improve the performance of DLGM by taking it as a QAP solver in our framework.

	\subsubsection{Spair-71K}

	The SPair-71K \cite{Spair71k} dataset contains 70,958 image pairs prepared from  PASCAL-VOC  2012  and  PASCAL  3D+.
	For our model, we adopt the same model architecture setting and training protocol as used on PASCAL-VOC  Keypoints.
	DGMC \cite{DGMC}, BBGM \cite{blackbox}, DLGM \cite{DLGM} and SIGMA \cite{SIGMA} are evaluated as comparison.
	The results in Table \ref{tb:spair} show that our model consistently improves upon the baseline, the one without ensemble, and out-performances DGMC, BBGM and SIGMA, while is slightly lower than DLGM.

\begin{table*}[tp]
\caption{\label{tb:few} Accuracies of few-shot 3D shape classification on the ModelNet40 dataset. Results of SPH \cite{few-shot-SPH}, LFD \cite{few-shot-LFD}, FV  \cite{few-shot-FV}, MVCNN \cite{few-shot-MVCNN}, GIFT\cite{few-shot-GIFT} and Meta-LSTM  \cite{few-shot-Neurocomputing} are taken from \cite{few-shot-Neurocomputing}. MFSC's results are taken from \cite{few-shot-Berkeley}. OOM indicates that the network caused a memory overflow.}
\begin{center}
\resizebox{0.995\textwidth}{!}{\begin{tabular}{cccccccccccc}
\toprule[2pt]
%~ &\multicolumn{3}{c|}{3D case} &\multicolumn{2}{c}{2D case} \\
%\bottomrule[0.1pt]
%\toprule[0.1pt]
Method  & SPH \cite{few-shot-SPH} & LFD \cite{few-shot-LFD} & FV  \cite{few-shot-FV}	& MVCNN \cite{few-shot-MVCNN} & GIFT\cite{few-shot-GIFT} & Meta-LSTM  \cite{few-shot-Neurocomputing}	& MFSC  \cite{few-shot-Berkeley} & ${\rm \textbf{EQAN}} $ & ${\rm \textbf{EQAN-U}}$ & ${\rm \textbf{EQAN-R}}$ \\
\midrule[0.1pt]
5-way-1-shot & 28.86 & 41.08 & 43.13	 & 44.90 & 43.25 & 45.46	& \textbf{68.10} & OOM & OOM & \textbf{66.24 $\pm 0.7$} \\
5-way-5-shot & 49.79 & 51.04 & 55.96 & 58.94 & 53.40 & 62.57 & 73.20 & OOM & OOM & \textbf{84.71 $\pm 1.4$} \\
\bottomrule[2pt]
\end{tabular}}
\end{center}
\end{table*}

\subsection{Few-shot 3D shape classification}

	We apply our model to the few-shot 3D shape classification task via point-cloud matching. 
	The experiment was conducted on the ModelNet40 dataset \cite{Modelnet40}, which consists of 12,311 meshed CAD models from 40 categories. In this dataset, the number of nodes in 3D shape objects is as large as thousands, causing computation challenges in graph matching. For few-shot classification, only 1 or 5 examples are used for each class for K-nearest neighbor classification.
	The selected classes are randomly sampled from 40 categories. We report the average accuracy of 20 independent runs.
	
    We compare our method with representative previous methods: SPH \cite{few-shot-SPH} uses the Spherical Harmonics descriptor for shape classification; LFD \cite{few-shot-LFD} uses the Light Field descriptor for 3D shapes; FV \cite{few-shot-FV} uses the Fisher vectors for the 3D object;  MVCNN \cite{few-shot-MVCNN} and GIFT \cite{few-shot-GIFT} adopt a CNN network to extract multi-view semantic features for a 3D object into multiple views; Meta-LSTM \cite{few-shot-Neurocomputing} uses a dual LSTM model to tackle the 3D few-shot problem; MFSC \cite{few-shot-Berkeley} is a model-agnostic approach for few-shot shape recognition.

	Following the  settings in \cite{few-shot-Berkeley,few-shot-Neurocomputing}, we report the results of two scenarios: the 5-way-1-shot case and the 5-way-5-shot case, to classify 5 unseen classes. 
	To test the models, we obtain the point cloud by uniformly sampling 1000 points from the surface  of each CAD model. We normalize those coordinates by subtracting the statistic mean and dividing by its standard deviation. Then, we connect each point in the point cloud with their top-3 nearest neighbors.

	We train our model using only 3D synthetic random graphs and only use the normalized coordinate information of keypoints as input node features. For constructing the ensemble model, the block number is set to be $L = 10$ and the solver number is $C = 6$ for each block.
	In addition, we use the random sampling mechanism introduced in Section \ref{sec:dis2} to reduce GPU memory usage.  The classification is based on the similarity $\bm{r}$ between  the query graph and the support graph.
	Let the matrix $\bm{R}$ be the reward matrix defined in  \eqref{reward_matrix}, and $\bm{Q}$ be the output matching prediction defined in  \eqref{matching_score}. The similarity $\bm{r}$ is computed via $ \bm{r} =  \optr(  \bm{Q} \bm{R}  ) $, i.e., the inner product between $\mQ$ and $\mR$.

	The classification results are shown in Table \ref{tb:few}. In the single-shot setting, our technique (EQAN-R) and MFSC \cite{few-shot-Berkeley} surpass all other methods by 20 percent. In the 5-shot circumstances, our technique  (EQAN-R) outperformed the other methods by more than 10 percent, while the vanilla version (EQAN) and the model with affinity update (EQAN-U) cannot handle such large 3D point-cloud graphs.

\subsection{Ablation study}
\label{sec:effect_design}
\revised{
We conduct a thorough ablation study to investigate which part makes the proposed ensemble-analogy GNN effective. 
Specifically,  we first investigate how to fuse the ensemble of base solvers, where the naive average only yields matching accuracy of  37.0, and our EQAN with special designed connectivity reaches 99.1.
Then, we study the effect of training the internal parameters in base solvers.
Next, we show the difference among 4 types of base solvers.
Additionally, we discuss the choice of input for making the final matching decision, and the effect of normalization layers. Finally, we study hyper-parameter sensitivity, random sampling strategy and real running time.}
	
	The experiments are based on 2D synthetic graphs randomly generated with  the number of inliers $n_\text{in}=35$ and number of outliers $n_\text{out} = 15$, noise level $\sigma = 0.3$ and we use the k-neighbor nearest rule  with $k=3$ to generate the edges.

	\subsubsection{Our ensemble model v.s. naive ensemble average}
	\label{sec:other_ensemble}
	In this experiment, we compare our ensemble model and the naive ensemble method. Our model inherits the idea of ensemble learning with a more involved structure, e.g. adding a $1\times 1$ convolution to promote the information exchange among channels. The naive ensemble method, on the other hand, computes the matching estimation independently, and only performs the average over the very last output.

	Specifically, we test a naive implementation of an ensemble model that the decision matrix generated by $\bm{Q} = \sum_{i} w_i \bm{z}_i$, where $\bm{z}_i$ is the $i$th independent solver's prediction, and $w_i$ is a learnable weight.
	The experiment listed in Table \ref{tb:ablation}(a) shows that the naive ensemble method does not gain any performance improvement compared to the single solver (denoted by w/o ensemble). This experiment indicates that the design of the ensemble architecture plays a crucial role.

    \subsubsection{{Learn or fix the internal parameters $\bm{w}$}}
    The parameters $\bm{w}$ inside the proximal solver determine the optimization trajectory and convergence of the solver, as illustrated in Proposition 1 (in the Appendix).
    Manually determining the best parameters for all solvers in the network is nearly impossible.
    In our framework, the parameters are learned in an end-to-end way. In this experiment, we also implement a variant model in which the internal parameters are determined by random sampling between 0 and 1. As shown in Table \ref{tb:ablation}(b), the learning model outperforms the random selection model  demonstrating the necessity of learning the internal parameters.

\begin{table}[tp]
\caption{\label{tb:ablation} Study the effect of architecture design on synthetic random graphs, where the underlined choices are adopted in our previous experiments. }
\begin{center}
\resizebox{0.4795\textwidth}{!}{\begin{tabular}{llc}
\toprule[1pt]
Component  & Choice  & Accuracy  \\
\midrule[0.1pt]
					\multirow{3}{*}{(a). Ensemble framework}
					&w/o ensemble & 37.4 \\
					~&Averaging & 37.0\\
					~ &\underline{EQAN} & 99.1\\
					\midrule[0.1pt]
					\multirow{2}{*}{\tabincell{c}{(b). Internal parameters $\bm{w}$ } }
					&Random &74.1 \\
					~ &\underline{Learning} & 99.1 \\
					\midrule[0.1pt]
					\multirow{8}{*}{(c). Ensemble with other solvers  }
					&DPGM-Single &37.1 \\
					&DPGM-Ensemble &99.1 \\
					&GAGM-Single & 39.9 \\
					&{GAGM-Ensemble} & {96.2} \\
					~ &SM-Single &38.6 \\
					~ &{SM-Ensemble} &94.7 \\
     &\revised{RRWM-Single} &\revised{40.1} \\
					&\revised{RRWM-Ensemble} &\revised{99.4} \\
     \midrule[0.1pt]
					\multirow{2}{*}{(d). Decision feature $\bm{V}$}
					&Last one only & 93.4 \\
					~  &\underline{All blocks} & 99.1\\
					
     \midrule[0.1pt]
\multirow{4}{*}{(e). Model width $C$    }
					&$C = 8$ &79.1  \\
					~ &$C = 16$ &89.2 \\
					~ &\underline{$C = 32$} &99.1 \\
					~ &$C = 64$ &99.3 \\
					\midrule[0.1pt]
					\multirow{5}{*}{(f).  Model depth  $L$ }
					&$L = 2$ &79.2 \\
					~&$L = 4$ &98.1  \\
					~&$\underline{L = 5} $ &99.1  \\
					~&$L = 8$ &99.2  \\
					~&$L = 16$ &98.9  \\
     \bottomrule[1pt]
			\end{tabular}}
		\end{center}
	\end{table}

	\subsubsection{Replace DPGM with other QAP-solvers}
	\label{sec:ensemble-other}
	In this experiment, we test different QAP-solver in our ensemble model, namely the differentiable graduated assignment method (GAGM) \cite{GAGM}, the differentiable (power-iteration based) spectrum method (SM) \cite{gmn} and RRWM \cite{RRWM}.
	Specifically, we use them to construct an EQAN with $L=5$ blocks, where each block contains 32 GAGM or SM solvers.
	For GAGM, by following the annealing strategy in \cite{GAGM}, we set the internal parameter $\beta$ for solver in the $\ell$th block to be $\beta = 0.5*1.075^{\ell-1}$.
	These  QAP-solvers can also be viewed as GNNs on the association graph. 
	We put their pseudo-code in Appendix B for the readers' convenience.
	The results shown in Table \ref{tb:ablation}(c) demonstrate that the proposed  ensemble model can significantly improve performance for all selected QAP solvers, which verifies the universality of our framework.

 \revised{
 Interesting, EQAN with the base solver of RRWM have slightly better performance (99.4) than the one with DPGM (99.1).  Compared with RRWM, DPGM and RRWM share similar unconstrained gradient update and the Sinkhorn  operation. However, RRWM has additional normalization and momentum term, which may contribute to the additional improvement.}

\subsubsection{The decision feature: all blocks or last one only}
	In our model, the decision layer takes the feature tensor $\bm{V}$ defined in \eqref{eq:concat} as the input feature. Originally, we concatenate output of all the EnsembleBlock. Here, we compare this decision feature with an alternative that only use the output of the last block $\bm{V} = \bm{V}^{(L)}$. Details are shown in Table \ref{tb:ablation}(d). The experiment verifies that the model using all the features of all blocks is better than the one just using the last layer's output.

\revised{
 \subsubsection{Sinkhorn normalization layer}
To study the effect of the Sinkhorn operation after each layer of GNN, we establish an experiment, in which we revised the GNN structure that only run a single Sinkhorn procedure before the decision layer. The accuracy decreases from the original 99.1\% to 71.4\% under the same setup as other ablation study. It indicates the Sinkhorn normalization plays an important role in the proposed GNN model. An intuitive explanation is the analogy of the Sinkhorn procedure as a (batch) normalization layer, which is a common module in convolutional neural networks, whereas the Sinkhorn procedure implements a more strictly, doubly stochastic normalization. 
}

 \subsubsection{Hyper-parameters: width and depth}
	Next, we study how different width ($C$, the number of solvers in each block) and depth ($L$, the number of blocks in the model) affect the results. We test the width $C$ from 8 to 64 while keeping the depth $L = 5$. Then, fixing the width $C=32$, we change the depth $L$ from 2 to 16. Experiments reveal that  the width and depth have an optimal region, and within this region, the performance is stable. Details are shown in Table \ref{tb:ablation} component (e) and (f). 
 In general, larger model results better performance, and the setting $C=32$ and $L=5$ is the best settings for balancing performance and model complexity.

	\subsubsection{The random sampling strategy}
\begin{figure}[tp]
		\centering
		\includegraphics[width=.898\linewidth]{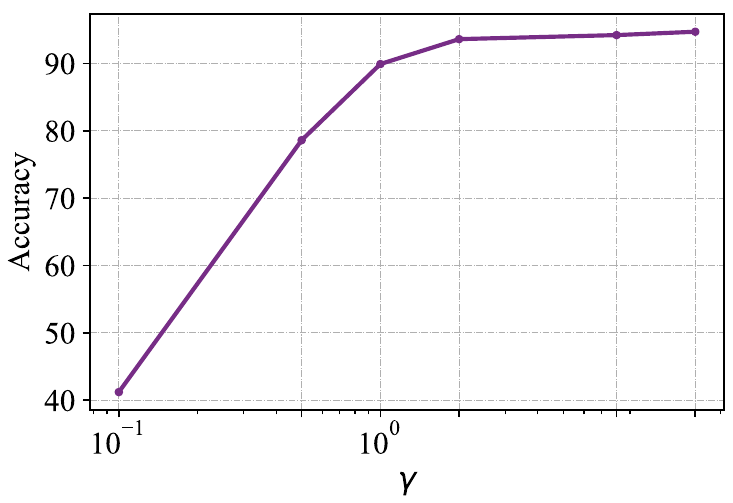}
		\caption{ \label{random} Random sampling size v.s. matching accuracy. 
	The graph matching accuracy under different sampling sizes  $\gamma N_s$ with ${\text{N}_{s} } = { n\sqrt{n}} = {5400}$.
	}
	\end{figure}

	We conduct experiments to study the relation between
	 the sampling size (number of sampled entries) and the matching accuracy. The random sampling strategy can reduce the GPU memory usage so that our model can scale up to thousand nodes. On the other hand, the random sampling trick can potentially affect the prediction accuracy.
	%In practice, the GPU memory footprint limits the model's scalability. The high-dimensional matrix-vector optimization in \eqref{eq:iter-1}, in particular, dominates the computational complexity of the QAP-solver. The random sampling technique approximates it by updating only a random subset of elements in \eqref{eq:iter-1} and directly copying the remaining entries from their corresponding input entries, lowering the GPU memory footprint complexity to an acceptable level.
	%(2) the inference speed with/without the sampling strategy;
	A moderate sampling size is important to balance the efficiency and the accuracy. We test different sampling size $\text{N}_{s} = \gamma n\sqrt{n}$ by changing the coefficient $\gamma$, where $n$ is the number of nodes of the input graph.
	The experimental results are reported in Figure \ref{random}.
	It shows that when $\gamma>1$, the performance will almost saturate. Larger $\gamma$
does not lead to significant performance gains, but does increase the model complexity.
	Empirically, $\gamma=1$ achieves the best trade-off between efficiency and accuracy.
 
	\minorrevised{In Table \ref{tb:random sample}, we  provide the comparison between our guided sampling scheme and the randomly sampling on PASCAL-PF. The results show that the proposed sampling scheme is better than the naive random sampling. }

\begin{table}[http]
\caption{\label{tb:random sample} \minorrevised{The EQAN  with  the proposed sampling scheme and the vanilla random sampling scheme. }}
\begin{center}
\resizebox{0.30395\textwidth}{!}{\begin{tabular}{llc}
\toprule[1pt]
 Choice  & Accuracy  \\
\midrule[0.1pt]
     %\cline{2-3}
     \midrule[0.1pt]
					Full messages (no sampling) \cite{GAGM} & 96.2 \\
					Proposed sampling scheme & 90.3 \\
					Vanilla random sampling scheme & {81.9} \\
					\bottomrule[1pt]
			\end{tabular}}
		\end{center}
\end{table}
 
	\subsubsection{\revised{Running time for inference}}
\begin{figure}[t]
		\centering
		\includegraphics[width=.998\linewidth]{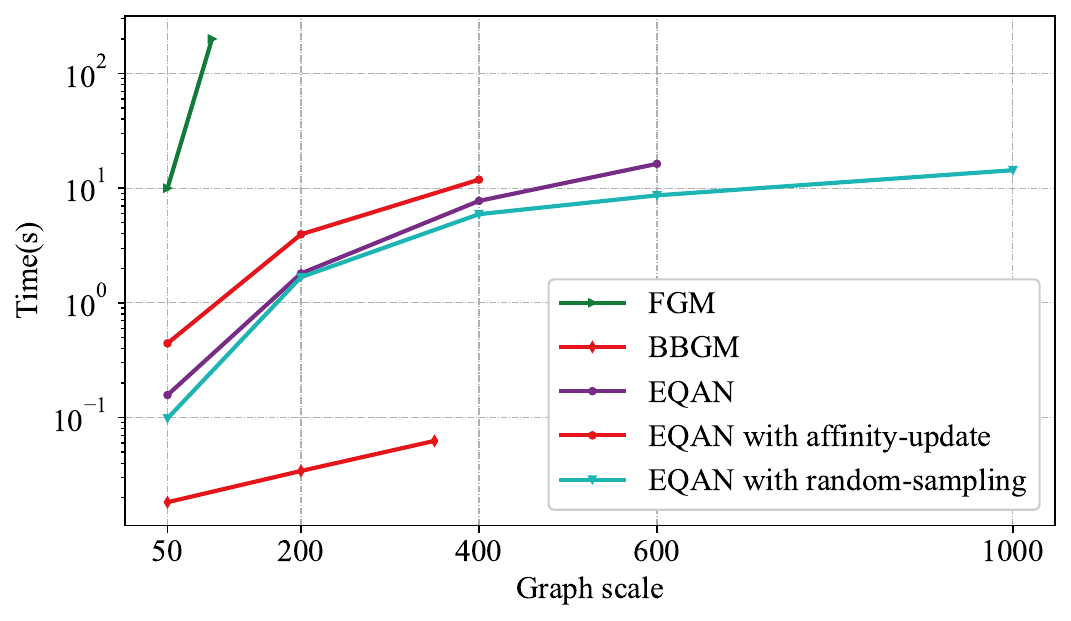}

		\caption{ \label{speed} Study on the random sampling strategy. 
	The inference running time of the vanilla network (EQAN), the network with the affinity update scheme (EQAN-U), the network with the random sampling strategy (EQAN-R) \revised{as well as two peer methods FGM \cite{fgm} and BBGM \cite{blackbox} }under different graph scales.
	}
\end{figure}

 	\revised{
	Figure \ref{speed} shows the real running time of the vanilla EQAN, the EQAN with the affinity update scheme (EQAN-U), and the EQAN with the random sampling strategy (EQAN-R) as well as two peer methods, FGM \cite{fgm} and BBGM \cite{blackbox}, under different graph scales. 
    Under the small graph setting ($N\leq 150$), the running time of three varieties of the proposed EQAN is less than one second, which is slower than the BBGM but significantly faster than the classic FGM algorithm.
    }    
	The vanilla EQAN model reaches the computational limit with more than 600 nodes.
	The affinity update scheme consumes more time,
	whereas the random sampling strategy allows the model to scale for the graphs up to 1000 nodes in a reasonable amount of time (about 15 seconds).
	We also note that when the graph size is small, the random sampling strategy does not significantly improve the inference speed of the model especially, as the sampling mask ${B}$ breaks the vectorized parallel computation at the hardware implementation level.

\section{Conclusion}

This paper proposes a new graph neural network (GNN) that combines the advantage of both traditional graph matching algorithms and recently proposed graph neural network (GNN) based approaches.
Specifically, we transform the previous traditional solvers as single-channel GNNs on the association graph and extend the single-channel architecture to the multi-channel network.
The proposed model can be seen as an ensemble method that fuses multiple algorithms at every iteration in a parameterized way.
In addition, we propose a random sampling strategy to reduce the computational complexity and GPU memory usage so as to improve the scalability of the model.
Experimental results of geometric graph matching, semantic feature matching, and few-shot 3D shape classification demonstrate the superior performance, robustness, and scalability of our approach.

\section*{Acknowledgements}
This work has been supported by the Major Project for New Generation of AI under Grant No. 2018AAA0100400, the National Natural Science Foundation of China (NSFC) grants U20A20223, 61836014, 61721004, the Youth Innovation Promotion Association of CAS under Grant 2019141, and the Pioneer Hundred Talents Program of CAS under Grant Y9S9MS08 and Y9S9MS08.

\section*{Appendix}

\appendix

\section{Convergence analysis and derivation of the differentiable proximal graph matching algorithm} \label{sec:convergence}
	%\subsubsection{Theoretical analysis}
	In this section, we show that under a mild assumption, the proximal graph matching algorithm converges to a stationary point within a reasonable number of iterations. The main technique in this analysis follows  \cite{p_vi}, which studied a general variational inference problem using the proximal-gradient method.
	
	Before presenting the main proposition, we first show a technical lemma.

    \textbf{}

    \begin{lemma} \label{lem:KL}
    	There exist a constant $\alpha >0$ such that for all $\vz_{t+1}$, $\vz_t$ generated during the forward pass of the solver, we have
    	$$
    	(\vz_{t+1} - \vz_{t})^\T \nabla_{\vz_{t+1}} D(\vz_{t+1}, \vz_{t}) \geq \alpha \norm{\vz_{t+1} - \vz_t}^2,
    	$$
    	where $D$ is the KL-divergence.
    \end{lemma}

We note that the largest valid $\alpha$ satisfying
Lemma 1 is ${1}/{2}$. One can find the proofs of both this lemma and the proposition shown below in the appendix.
Next, we show the main result of our convergence analysis.

    \begin{proposition} \label{prop:conv}
    	Let $L$ be the Lipschitz constant of the gradient function $\nabla g(\vz)$, where $g(x)$ is defined in \eqref{eq:prox}, and let $\alpha$ be the constant used in Lemma~\ref{lem:KL}.
    	If we choose a constant step-size $\beta < 2\a/L$ for all $t\in \{0, 1, \ldots, T \} $, then
    	$$
    	\mathbb{E}_{t \sim \text{uniform} \{0, 1, \ldots, T \} }  \norm{\vz_{t+1} - \vz_t}^2
    	\leq \frac{C_0}{ T(\alpha - L \beta /2)},
    	$$		where $C_0 = |\mathcal{L}^\ast - \mathcal{L}(\vz_0)| $ is the objective gap of \eqref{eq:opt2} between the initial guess
    	$\mathcal{L}(\vz_0)$ and the optimal one $\mathcal{L}^\ast$.
    \end{proposition}
    Proposition~\ref{prop:conv} guarantees that the iterative process converges to a stationary point within a reasonable number of iterations.
    In particular, the average difference $ \norm{\vz_{t+1} - \vz_t}^2$ of the iterand converges with a rate
    $1/T$.

	\subsection{Proof of the Lemma \ref{lem:KL}}
	Because the proximal function $D(\vz, \vz_t)$ is convex, the following inequality always hold:
	\begin{equation*}\begin{split}
	D(\vz_t,\vz_t) \geq D(\vz_{t+1},\vz_t) + [\nabla_{\vz = \vz_{t+1}} D(\vz,\vz_t)]^\T (\vz_t - \vz_{t+1}).
	\end{split}\end{equation*}
	By using $D(\vz_t,\vz_t) = 0$, we obtain:
	\begin{equation*}\begin{split}
	D(\vz_{t+1},\vz_t) \leq  (\vz_{t+1} - \vz_t)^\T[\nabla_{\vz = \vz_{t+1}} D(\vz,\vz_t)].
	\end{split}\end{equation*}

	Let $l(\vz) = \vz^\T {\rm log} \vz$. We decompose $D(\vz_t,\vz_t)$ as the following form of bregman divergence:
	\begin{equation*}\begin{split}
	D(\vz_{t+1},\vz_t) =  l(\vz_{t+1}) - l(\vz_{t}) - \nabla l(\vz_{t})^\T(\vz_{t+1} - \vz_{t}).
	\end{split}\end{equation*}
	Moreover, due to the strong convexity of $l(\vz)$,
	\begin{equation*}\begin{split}
	l(\vz_{t+1}) - l(\vz_t) - \nabla l(\vz_t)^\T (\vz_{t+1} - \vz_{t}) \geq \frac{1}{2}||\vz_{t+1} - \vz_{t}||^2,
	\end{split}\end{equation*}
	then we get:
	\begin{equation*}\begin{split}
	\frac{1}{2}||\vz_{t+1} - \vz_{t}||^2 \leq D(\vz_{t+1}, \vz_t) \leq (\vz_{t+1} - \vz_{t})^T \nabla D(\vz_{t+1},\vz_{t}),
	\end{split}\end{equation*}
	which proves lemma 1 and suggests that the largest valid $\alpha$ is $\frac{1}{2}$.

\subsection{Proof of the Proposition \ref{prop:conv}}

		Before the proof of the proposition \ref{prop:conv}, we first present a technical lemma.
	 \begin{lemma} \label{lem:2} For any real-valued vector $\vg$ which has the same dimension of $\vz$ and $\beta>0$, considering the convex problem
    	\begin{equation}\begin{split}
    			\vz_{t+1} = &\mathop{\rm argmin}\limits_{\textbf{C}\vz = \textbf{1}}:~  \{ \vz^\T\textbf{g} - h(\vz) + \frac{1}{\beta} D(\vz,  \vz_t)     \},
    			\label{problem_g}
    	\end{split}\end{equation}
    	where $h(\vz) = -\lambda\vz^\T {\rm log }(\vz)$ and D is the KL-divergence, the following inequality always holds:
    	\begin{equation*}\begin{split}
    			\textbf{g}^\T(\vz_{t} - \vz_{t+1}) \geq \frac{\alpha}{\beta}||\vz_{t} - \vz_{t+1}||^2 - [h(\vz_{t+1}) - h(\vz_{t})].
    	\end{split}\end{equation*}
    \end{lemma}
    %One can show that the largest valid $\alpha$ satisfying \eqref{eq:in} is $\frac{1}{2}$.

	Because of the convexity of the objective in the sub-problem, it is easy to derive that, if $\vz^{*}$ is the optimal, the following hold:
	\begin{equation*}\begin{split}
	(\vz^{*} - \vz_t)^T(\textbf{g} - \nabla h(\vz^{*}) + \frac{1}{\beta} \nabla D(\vz^{*},  \vz_{t})) \leq 0 .
	\end{split}\end{equation*}
	Let $\vz^{*} = \vz_{t+1}$, by using lemma 1 we obtain that
	\begin{equation}\begin{split}
	\textbf{g}^\T(\vz_t - \vz_{t+1}) - (\vz_t - \vz_{t+1})^\T\nabla h(\vz_{t+1})
	- \frac{\alpha}{\beta}|| \vz_{t+1} - \vz_{t} ||^2  \\
	\geq 0. \label{ineq22}
	\end{split}\end{equation}
	Because the entropic function $h(\vz)$ is concave, hence
	\begin{equation}\begin{split}
	(\vz_t - \vz_{t+1})^\T\nabla h(\vz_{t+1}) \geq h(\vz_{t}) - h(\vz_{t+1}). \label{ineq23}
	\end{split}\end{equation}
	We can derive the following inequality from \eqref{ineq22} and \eqref{ineq23}:
	\begin{equation*}\begin{split}
	\textbf{g}^\T(\vz_t - \vz_{t+1}) - [ h(\vz_{t}) - h(\vz_{t+1}) ]
	- \frac{\alpha}{\beta}|| \vz_{t+1} - \vz_{t} ||^2  \\
	\geq 0,
	\end{split}\end{equation*}
	which proves lemma 2.

	Now, we start to prove Proposition \ref{prop:conv}.
	\begin{proof}:
		Because $g(\vz)$ is $L$-Lipschitz gradient continuous, we  get
		\begin{equation}\begin{split}
		g(\vz_{t+1}) \leq g(\vz_{t}) + \nabla g(\vz_{t})^\T (\vz_{t+1} - \vz_{t})
		+ \frac{L}{2} ||\vz_{t+1} - \vz_{t}||^2. \label{propo: tempsetp}
		\end{split}\end{equation}
		Let $\vg = \nabla g(\vz_{t})$. By using lemma 2, we  bound the right side of \eqref{propo: tempsetp} by
		\begin{equation}\begin{split}
		g(\vz_{t+1}) \leq g(\vz_{t}) - \frac{\alpha}{\beta}||\vz_t - \vz_{t+1}||^2 - [h(\vz_t) - h(\vz_{t+1})]\\
		+ \frac{L}{2} ||\vz_{t+1} - \vz_{t}||^2. \label{temp_ineq1}
		\end{split}\end{equation}
		Rearranging the terms in \eqref{temp_ineq1} and noting that $g(\vz) - h(\vz) = L(\vz)$, we get
		\begin{equation*}\begin{split}
		(\frac{\alpha}{\beta} - \frac{L}{2}) ||\vz_{t+1} - \vz_{t}||^2 \leq  \mathcal{L}(\vz_{t}) - \mathcal{L}(\vz_{t+1}).
		\end{split}\end{equation*}
		Next, we choose a constant step-size $\beta < 2\a/L$ for all $t\in \{0, 1, \ldots, T \} $. By summing both side from index of $0$ to $T$, we have
		\begin{equation*}\begin{split}
		\frac{1}{T}\sum_{t=0}^{T}(\frac{\alpha}{\beta} - \frac{L}{2}) ||\vz_{t+1} - \vz_{t}||^2 \leq  \frac{\mathcal{L}(\vz_{0}) - \mathcal{L}(\vz_{T})}{T}  \\
		\leq \frac{\mathcal{L}(\vz_{0}) - \mathcal{L}^\ast}{T},
		\end{split}\end{equation*}
		and finally reach
		\begin{equation*}\begin{split}
		\mathbb{E}_{t \sim \text{uniform} \{0, 1, \ldots, T \} }  \norm{\vz_{t+1} - \vz_t}^2 \leq  \frac{ \mathcal{L}(\vz_{0}) - \mathcal{L}^\ast }{ T(\alpha - L \beta /2) }.
		\end{split}\end{equation*}
	\end{proof}

\subsection{\revised{Derivation of DPGM Algorithm}}
\label{Derivation}

\revised{
Our proximal graph matching solves a sequence of convex optimization problems
\begin{equation*} \label{eq:prox-GM} \scriptsize
\begin{aligned}
\vec{z}_{t+1} =& \mathop{\rm argmin}_{\vz \in \R_{+}^{n^2\times 1}} - \vz^\T (\vu + \mP \vz_t)+ \tfrac{1+ \beta_t}{\beta_t}  \vz^\T \log (\vz) 
- \tfrac{1}{\beta_t} \vz^\T \log(\vz_t)
\\
\text{s.t. }\quad & \mC \vz = \mat{1}.
\end{aligned}
\end{equation*}
where the binary matrix $\mC \in \{0,1\}^{n^2 \times n^2}$ encodes $n^2$ linear constraints ensuring that
$\sum_{i\in V_1} x_{ij^\prime} = 1 $ and $\sum_{j \in V_2} x_{i^\prime j}=1$ for all $i^\prime \in V_1$ and $j^\prime \in V_2$.
We set 
\begin{equation*} \label{eq:prox-GM}
\begin{aligned}
E_t(\vz) = - \vz^\T (\vu + \mP \vz_t)+  \tfrac{1+\beta_t}{\beta_t} \vz^\T \log (\vz) - \tfrac{1}{\beta_t} \vz^\T \log(\vz_t)
\end{aligned}
\end{equation*}
The matching objective with Lagrange multipliers for each sub-problem is
\begin{equation*}\begin{split}
\mathcal{L}(\vz, \mat{\mu}, \mat{\nu}) =    E_t(\vz) + \mat{\mu}^\T(\mat{Z}\mat{1} - \mat{1})  + \mat{{\nu}}^\T(\mat{Z}^\T\mat{1} - \mat{1}), \label{eq:lagrange}
\end{split}\end{equation*}
where $\mat{\mu}$ and $\mat{\nu}$ are Lagrange multipliers, and  $\mat{Z}$ is the matrix form of vector $\vz$, which should be a doubly stochastic matrix. 
Setting derivatives  
$\frac{\partial{\mathcal{L}}}{\partial \vz}$,  $\frac{\partial{\mathcal{L}}}{\partial \mat{\mu}}$, $\frac{\partial{\mathcal{L}}}{\partial \mat{\nu}}$ be 0, we get
\begin{align*}
&[\vz]_{ij} = \exp\Big[ \tfrac{\beta_t}{ 1+ \beta_t} [ \vu + \mP \vz_t]_{ij} 
+ \tfrac{1}{1+\beta_t} \log([\vz_t]_{ij})
\\& \quad \quad+ 1 +\tfrac{\beta_t}{ 1+ \beta_t}\big( [\mat{\mu}]_j +  [\mat{\nu}]_i \big)\Big]
\\
&\sum_{i^\prime = 1}^n [\vz]_{i^\prime} = 1, \sum_{j^\prime = 1}^n [\vz]_{ij^\prime } = 1,
\end{align*}
for all $ i,j \in 1,2,\ldots, n$.
The solution of the above equations yields the following update rule \cite{softmax_to_softassignment,cccp}
\begin{align} \label{eq:iter-1a}
\widetilde{\vz}_{t+1} &=
\exp\Big[
\tfrac{\beta_t}{ 1+ \beta_t} ( \vu + \mP \vz_t) 
+ \tfrac{1}{1+\beta_t} \log(\vz_t)
\Big]
\\ \label{eq:iter-2a}
\vz_{t+1} & = \text{Sinkhorn}(\widetilde{\vz}_{t+1}),
\end{align}
where $\text{Sinkhorn}(\vz)$ is the Sinkhorn-Knopp transform \cite{sinkhorn} that  maps 
a nonnegative matrix of size $n\times n$ to a doubly stochastic matrix.
Here, the input and output variables are $n^2$ vectors. When using the Sinkhorn-Knopp transform, we reshape the input (output) as an $n\times n$ matrix ( $n^2$-dimensional vector) respectively.
}

\revised{
Given a nonnegative matrix $\boldsymbol{F} \in \mathbb{R}_{+}^{n\times n}$, Sinkhorn algorithm works iteratively. In each iteration, it normalizes all its rows via the following equation:
\begin{equation*}\begin{split}
\boldsymbol{F}^{'}_{ij} = \boldsymbol{F}^{'}_{ij}/(\sum_{k=1}^{n} \boldsymbol{F}_{ik}),
\end{split}\end{equation*}
then it takes the column normalization by the following rule:
\begin{equation*}\begin{split}
\boldsymbol{F}_{ij} = \boldsymbol{F}_{ij}/(\sum_{k=1}^{n} \boldsymbol{F}^{'}_{kj}),
\end{split}\end{equation*}
After processing iteratively until convergence, the original matrix $\boldsymbol{F}$ would be transformed into a doubly stochastic matrix. 
Equation \eqref{eq:iter-1a} and \eqref{eq:iter-2a} lead to the proximal iteration \eqref{eq:iter-1} and \eqref{eq:iter-2}. 
}

\section{Additional pseudo-code of QAP solvers}
\label{appendix2}
We presented the pseudocodes of the related classical matching algorithms, GAGM and spectral method (SM) for readers' reference.
Details are shown in Algorithm~\ref{alg:GAGM} and \ref{alg:SM} respectively.

\begin{algorithm}[http]
	%\large
	\caption{:~Graduated assignment (GAGM) \label{alg:GAGM}}
	\begin{algorithmic}
		\STATE {\bfseries Input:}  Affinity matrix $\bm{M}$, maximum iteration $K$, the annealing factor $\{ \beta_1, ..., \beta_{K} \}$.
		%%%%%%%%%%%%%%%%%%%%%%%%%%%%%%%%%%%%%%%%%%%%%%%%%%%%%%%%%%%%%%
		\STATE {\bfseries Initialization:} as the uniform starting solution $\bm{x}_0$.
		%%%%%%%%%%%%%%%%%%%%%%%%%%%%%%%%%%%%%%%%%%%%%%%%%%%%%%%%%%%%%%
		\FOR{$k=1$ {\bfseries to} $K$}
		\STATE \quad $\hat{\bm{x}}_{k} = \underbrace{\bm{M}\bm{x}_{k-1}}_{\text{Message passing}}$
		\STATE \quad $\underbrace{  \bm{x}_{k} =    \text{Sinkhorn}(\exp (\beta_k \hat{\bm{x}}_{k})   )        }_{\text{Normalization}}$
		%%%%%%%%%%%%%%%%%%%%%%%%%%%%%%%%%%%%%%%%%%%%%%%%%%%%%%%%%%%%%%
		\ENDFOR
		\STATE {\bfseries Output:} $\bm{x}_{K}$
	\end{algorithmic}
\end{algorithm}

\begin{algorithm}[http]
	%\large
	\caption{:~Power-method based spectral method (SM) \label{alg:SM}}
	\begin{algorithmic}
		\STATE {\bfseries Input:}  Affinity matrix $\bm{M}$, and maximum iteration $K$.
		\STATE {\bfseries Initialization:}  the uniform solution $x_{0}$.
		\FOR{$k=1$ {\bfseries to} $K$}
		\STATE \quad $\hat{\bm{x}}_{k} = \underbrace{\bm{M} \bm{x}_{k-1}}_{\text{Message passing}}$
		\STATE \quad $\bm{x}_{k} = \underbrace{ \hat{\bm{x}}_{k}/\norm{\hat{\bm{x}}_{k}} }_{\text{Normalization}}$
		\ENDFOR
		\STATE {\bfseries Output:} $\bm{x}_{K}$
	\end{algorithmic}
\end{algorithm}

%\begin{acknowledgements}
%\end{acknowledgements}



\bibliographystyle{spmpsci}
\bibliography{main.bib}

\end{document}